\documentclass[journal]{IEEEtran}
\usepackage{bm}
\usepackage{amsmath}
\usepackage{amsfonts}
\usepackage{graphicx}
\usepackage{algorithm}
\usepackage{algorithmic}
\usepackage{multirow}
\usepackage{color}
\usepackage{subcaption}
\DeclareMathOperator*{\Diag}{Diag}
\DeclareMathOperator*{\Tr}{Tr}
\DeclareMathOperator*{\x}{\mathbf{x}}
\DeclareMathOperator*{\X}{\mathbf{X}}
\DeclareMathOperator*{\y}{\mathbf{y}}
\DeclareMathOperator*{\Y}{\mathbf{Y}}
\DeclareMathOperator*{\A}{\mathbf{A}}
\DeclareMathOperator*{\J}{\mathbf{J}}
\DeclareMathOperator*{\e}{\mathbf{e}}
\DeclareMathOperator*{\M}{\mathbf{M}}
\DeclareMathOperator*{\I}{\mathbf{I}}
\DeclareMathOperator*{\vv}{\mathbf{v}}
\DeclareMathOperator*{\rr}{\mathbf{r}}
\ifCLASSINFOpdf
\else
\fi

\graphicspath{{./Figure/}}
\begin{document}
%
\title{Robust Face Recognition via Adaptive Sparse Representation}


\author{Jing Wang, Canyi Lu, Meng Wang,~\IEEEmembership{Member,~IEEE,} Peipei Li,\\ 
Shuicheng Yan, ~\IEEEmembership{Senior Member,~IEEE,} Xuegang Hu
\thanks{J. Wang, M. Wang, P. Li and X. Hu are with the School of Computer Science and Information Engineering, Hefei University of Technology, Hefei 230009, China (e-mail: hfutwj@gmail.com; eric.mengwang@gmail.com; peipeili.hfut@gmail.com; jsjxhuxg@hfut.edu.cn).}
\thanks{C. Lu and S. Yan are with the Department of Electrical and Computer Engineering at National University of Singapore, Singapore (e-mail:canyilu@gmail.com; eleyans@nus.edu.sg)}
}

%



\IEEEtitleabstractindextext{
\begin{abstract}
Sparse Representation (or coding) based Classification (SRC) has gained great success in face recognition in recent years. However, SRC emphasizes the sparsity too much and overlooks the correlation information which has been demonstrated to be critical in real-world face recognition problems. Besides, some work considers the correlation but overlooks the discriminative ability of sparsity. Different from these existing techniques, in this paper, we propose a framework called Adaptive Sparse Representation based Classification (ASRC) in which sparsity and correlation are jointly considered. Specifically, when the samples are of low correlation, ASRC selects the most discriminative samples for representation, like SRC; when the training samples are highly correlated, ASRC selects most of the correlated and discriminative samples for representation, rather than choosing some related samples randomly. In general, the representation model is adaptive to the correlation structure, which benefits from both $\ell_1$-norm and $\ell_2$-norm. 
   Extensive experiments conducted on publicly available data sets verify the effectiveness and robustness of the proposed algorithm by comparing it with state-of-the-art methods.
\end{abstract}

}

\maketitle

\IEEEdisplaynontitleabstractindextext
\IEEEpeerreviewmaketitle

\section{Introduction}
\IEEEPARstart{F}{ace} recognition has drawn intensive interest in pattern recognition for decades due to its wide real-world applications, such as video surveillance, person tracking, and access control \cite{zhao2003face}\cite{ortiz2013movie}\cite{ocegueda20133d}. However, the images taken by the devices under the unconstrained environment are usually of limited quality. Various human facial expressions, poses, and illumination conditions affect the quality of face images, causing occlusion, translation and scale errors, etc. in the normalized face images, as shown in Figure 1. Furthermore, handling these problems in the high dimensional feature space or on an under-sampled dataset makes the task of face recognition even more challenging. Therefore, although some recognition methods have been proposed, and obtained great success during the past few years, robust face recognition methods with higher recognition performances are still desired.

 One major type of face recognition methods adopt the holistic model, which identifies the label of the image by global representation \cite{brunelli1993face}. These methods are also called appearance-based methods. The other type of methods utilize the component based model which first divides the face image into patches and then extracts features from each patch \cite{heisele2003face}. Finally the classification decision is made based on the similarity between patches of different images. These methods usually adopt manifold learning \cite{kokiopoulou2007orthogonal} and are suitable for under-sample problems \cite{lu2011discriminative}. Some other researchers address the problem by feature extraction. Besides the classical features such as Eigenfaces \cite{eign} and Gabor \cite{daugman1985uncertainty}, some robust features are also proposed. For instance, Gradientfaces is an illumination insensitive measure and robust to various illuminations \cite{zhang2009face}. The subspace learning from gradient orientations was demonstrated to be robust to different noises for objective recognition \cite{tzimiropoulos2012subspace}. Chan et al. \cite{chan2013multiscale} proposed a novel descriptor based on Local Phase Quantization (LPQ) which is insensitive to blur. They extended the descriptor to a multiscale framework combined with Multiscale Local Binary Pattern to increase the insensitivity to illumination. Based on the feature space, existing classical classifiers can be used, such as NN (Nearest Neighbor), Support Vector Machines (SVM) \cite{CC01a} and boosting \cite{freund1999short}. NN belongs to the Nearest Feature based Classifiers (NFCs) which are based on the distance measure and have attracted much attention. However, the performance of NFCs may degrade sharply when the images of different classes are quite similar.

 To overcome these limitations, sparse representation is employed in face recognition. Sparse Representation based Classification (SRC) \cite{robust} seeks a representation of the query image in terms of the over-complete dictionary
, and then performs the recognition by checking which class yields the least representation error. Thus SRC can be considered as a generalization of NN and NFS~\cite{shan2002face}, but it is more robust to occlusions and variations. \cite{robust} reported that SRC remains $100 \%$ recognition rate for even $60\%$ randomly corrupted pixels. The striking performance of SRC boosts the research of sparse representation based face recognition and brings a promising research direction. For example, Elhamifar and Vidal \cite{elhamifar2011robust} proposed a structured sparse representation, while Gao et al. \cite{gao2010kernel} introduced a kernel sparse representation. Cheng et al. \cite{cheng2010l1} introduced the $\ell_1$-graph for image analysis, and Yang et al. \cite{yang2009linear} integrated sparse coding with linear spatial pyramid matching for image classification. Deng et al. \cite{deng2012extended} introduced an intraclass variant dictionary into SRC for undersampled face recognition, and Oritiz \cite{ortiz2013movie} proposed mean sequence sparse representation based classification for face recognition in movie trailers.
Tzimiropoulos et al.
\cite{tzimiropoulos2011sparse} proposed sparse representation based on image gradient orientations for visual recognition and tracking.
 Unfortunately, SRC places too much emphasis on sparsity and overlooks the correlation within the dictionary which will lead to information loss. Especially, SRC will produce unstable results when the training samples are highly correlated. Some works have demonstrate the importance of correlation structure \cite{wang2010locality}\cite{lu2013image}\cite{wang2013view}\cite{wang2012assistive}. Specifically, Zhang \cite{rigamonti2011sparse} proposed CRC (Correlation Representation based Classifier) in consideration of the collaborative representation. CRC employed $\ell_2$-norm to obtain a denser detector. However, the $\ell_2$-norm does not perform sample selection which tends to include training samples of various classes and disturbs recognition results in some ways.

To address the above problems, we propose an Adaptive Sparse Representation based Classification (ASRC) method inspired by the work of trace norm \cite{traceLasso}. Trace norm captures the correlations among multiple variables and has been employed in many applications. Yang et al.~\cite{yang2013feature} used the trace norm to find the sharing information among multiple tasks for feature selection. In \cite{ma2012knowledge}, the concepts adaptation was employed to assist event detection. Similarly, ASRC employs the trace norm on the dictionary and the representation vector to form a correlation adapter, which considers both correlation and sparsity. Our model is built to find a representation vector that minimizes the correlation regularizer. We also give theoretical analysis to prove that the regularizer balances the $\ell_1$-norm and $\ell_2$-norm with adaptive consideration of data structure. Thus our scheme can obtain a more accurate representation with the optimal discriminative and correlated training samples. 
To sum up, our major contributions are as follows:

\begin{itemize}
\item In this paper, we propose an adaptive sparse linear model for face recognition with joint consideration of correlation and sparsity. Based on the model, a face recognition method ASRC is presented and it is adaptive to the exact structure of the dictionary.

 When the training samples are barely correlated, ASRC acts like SRC. When the training samples are highly correlated, ASRC is equivalent to CRC. In general, the sparsity of the representation vector obtained by ASRC is between the ones obtained by SRC and CRC.

\item We perform extensive experiments on three publicly available face data sets and fifteen UCI data sets. The experimental results demonstrate that ASRC is superior to existing state-of-the-art face recognition methods, such as NN, NFS, SRC, CRC and LSRC.
\end{itemize}

\begin{figure}[!t]
\centering
\includegraphics[width=0.5\textwidth]{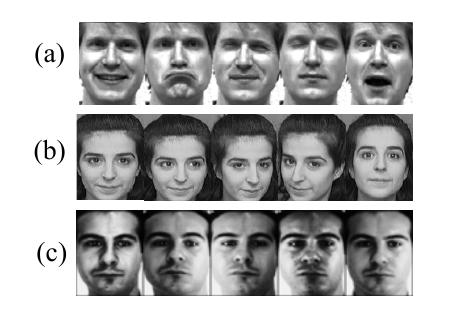}
\caption{Face image samples. (a) Face images with various expressions from the Yale database. (b) Face images with different poses from the ORL database. (c) Face images with different illuminations from the AR database.}
\label{fig_uci}
\end{figure}

The remainder of this paper is organized as follows. Section \ref{sec2} reviews some related work, including nearest feature classifiers and sparse coding based algorithms. Section \ref{sec3} introduces our algorithm and provides the solution for optimization. Extensive experimental results are presented and analyzed in Section \ref{sec4}. Finally, we conclude this paper in Section \ref{sec5}.

\section{Related Work}
\label{sec2}
In this section, we briefly introduce some works on the representation of images. We review the most popular face recognition methods, including Nearest Feature based Classifiers (NFCs) and sparse coding  based methods. We mainly discuss the original SRC approach. Before introducing these methods, we first describe the problem and explain the notations used in this work.

\textbf{Problem Description.} Denote the data set of training samples labeled with the $k$-th class as $\X_k=[\textbf{x}_{1}^k, \textbf{x}_{2}^k, \ldots, \textbf{x}_{n_{k}}^{k}]\in R^{m\times n_{k}}$, where each element is a training sample, $m$ is the dimension of the feature space and $n_{k}$ is the total number of $\X_k$. Suppose the dictionary has $K$ classes of samples denoted as $\X=[\X_1, \X_2, \cdots, \X_K]\in R^{m\times n}$, and $n=\sum_{i=1}^K n_{i}$. Given a query sample $\y\in R^{m}$, the pattern recognition task is to determine which class $\y$ belongs to.

\textbf{Notations.} Scalars are denoted by lowercase letters, e.g. $a$. Vectors are denoted by boldface lowercase letters, e.g. $\vv$. Matrices are denoted by boldface capital letters, e.g. $\M$.
 Especially, $\I$ denotes the identity matrix. For a vector $\vv$, $\Diag(\vv)$ indicates converting the vector $\vv$ into a diagonal matrix in which the $i$-th diagonal entry is $v_i$. The various vector norms we use in this paper are as follows:
\begin{itemize}
\item $|| \vv || _0$ is the $\ell_0$-norm, $i.e.$, the number of non-zero entries in the vector $\vv$.

\item $||\vv||_1$ is the $\ell_1$-norm, $i.e.$, the sum of the absolute values of all the entries, defined as $||\vv||_1=\Sigma _{i}|v_{i}|$.

\item $||\vv||_2$ is the $\ell_2$-norm, defined as $||\vv||_2=\sqrt{\Sigma _{i}v_{i}^2}$.

\item $||\vv||_\infty$ is the infinite norm, defined as $||\vv||_\infty=\max_{1\leq i\leq n}|v_i|$.

\end{itemize}
For a matrix $\M$, $\text{diag}(\M)$ indicates converting the matrix $\M$ into a vector in which the $i$-th entry is $M_{ii}$, the diagonal entries of $\M$. The $\ell_1$-norm of $\M$ is defined as $||\M||_1=\Sigma _{ij}|m_{ij}|$, and $||\M||_{\ast}$ is the trace norm, $i.e.$, the sum of all the singular values of the matrix $\M$.

Generally speaking, Nearest Feature based Classifiers aim to find a representation of the query image, and classify it to the best representer. According to the mechanism of representing the query image, NFCs include Nearest Neighbor (NN), Nearest Feature Line (NFL) \cite{li1999face}, Nearest Feature Plane (NFP) \cite{chien2002discriminant} 
and Nearest Feature Subspace (NFS). More specifically, NN is the simplest one with no parameters which classifies the query image to its nearest neighbor. As NN adopts only one sample to represent the query image, its performance can be easily affected by noises. Li and Lu proposed the NFL classifier which forms a line by every two training samples of the same class and classifies the query image to its nearest line. Chien and Wu proposed the NFP classifier which uses at least three training samples of the same class to form a plane rather than a line to determine the label of the query image. NN, NFL and NFP all use a subset of the training samples with the same label to represent the query image, while NFS represents the query image by all the training samples of the same class. In general, the more samples are used for representation, the more stable a method is supposed to be. Hence, NFS is assumed to perform better than the other NFCs. 
Besides, NFCs are not robust in real-world face recognition applications because of various occlusions.

To overcome these limitations, Wright et al. introduced the Sparse Representation based Classification (SRC) method to represent the query image $\y$ based on the over-complete dictionary $\X$ with sparse coefficients,
\begin{equation}\label{L0}
\min{||\bm{\alpha}|| _{0}} ~~ s.t. \  \y = \X\bm{\alpha}.
\end{equation}
The above $\ell_0$-norm minimization problem is non-convex and actually NP-hard. It has been proved in \cite{donoho2006most}\cite{candes2006stable} that problem (\ref{L0}) is equal to $\ell_1$-minimization problem under certain conditions:
\begin{equation}\label{L1}
\min{||\bm{\alpha}|| _{1}} ~~ s.t. \  \y = \X\bm{\alpha}.
\end{equation}
To deal with the noises, the $\ell_1$-minimization problem is extended to the following formulation:
\begin{equation}\label{SRC}
\min{||\bm{\alpha}|| _{1}} ~~ s.t. \  \|\y - \X\bm{\alpha}\|_2 \leq \varepsilon,
\end{equation}
where $\varepsilon > 0$ is a given tolerance. It is generally regarded that SRC is an extension of NN and NFS. The difference is that the coding of $\y$ is performed over the over-complete dictionary $\X$ instead of its subset. With a sufficient number of training samples, SRC with random projection-based features can outperform NFCs on conventional features. The sparse model is also more robust and effective for object recognition in the case that objects are corrupted by outliers. 

However, SRC is under the assumption that the query image $\y$ and the training images are well aligned. It is indicated that with sufficient training samples that cover nearly all the possible variations, $\y$ can be correctly represented and robust to variations \cite{wagner2009towards}. Thus, SRC may fail in the case that the query image is misaligned or the dictionary has a small number of samples. Meanwhile, due to the sparsity, SRC may have the instability problem when samples are highly correlated. If the subjects correlated to the query image look similar, the SRC method tends to select one at random for representation rather than selecting them all. It indicates that SRC fails to capture the correlation structure of the dictionary which is critical in face recognition \cite{zou2005regularization}.

Some works have focused on the above problem and provided several solutions. Wang et al. \cite{wang2010locality} proposed locality constraints in spatial sparse representation.
Li et al. \cite{li2009maximizing} maximized the intra-individual correlations to address the pose difference for face recognition. Zhang et al. imposed the $\ell_2$-norm constraint on the coefficients and proposed CRC (Collaborative Representation based Classification with regularized least square). In \cite{zhang2011sparse}, Zhang pointed out that the success of SRC comes from the collaborative representation of $\y$ over all training samples. The $\ell_2$-norm is supposed to take advantage of data correlation \cite{LSR}. Thus, the query image $\y$ is represented by the over-complete dictionary with $\ell_2$-norm rather than $\ell_1$-norm to constrain the coding vector in CRC. The objective function is defined as:
\begin{equation}\label{CRC}
\min{||\bm{\alpha}|| _{2}} ~~ s.t. \  \|\y - \X\bm{\alpha}\|_2 \leq \varepsilon.
\end{equation}
The $\ell_2$-minimization guarantees that CRC gets a stable result with a much denser representation vector. However, CRC does not perform sample selection for representation. It may not perform well when the training data are not highly correlated.

To overcome the drawbacks of the above works, we propose a novel sparse representation method, called Adaptive Sparse Representation based Classification (ASRC). Rather than adopting the $\ell_1$-norm and $\ell_2$-norm, ASRC imposes the trace norm on the representation vector combined with the structure of the data matrix. We also impose the $\ell_1$-norm on the noise part which makes our method robust to occlusions in the images. The details of our ASRC are presented in the following section.

\section{Adaptive Sparse Representation based Classification}
\label{sec3}
In this section, we detail our adaptive representation model and provide the solution to the optimization. Then we describe our face recognition algorithm ASRC. By analyzing the properties of the ASRC, we show how to improve the performances of SRC and CRC with adaptive incorporation of both correlation and sparsity.

\subsection{Our Model}
Variations in facial expressions or views, as well as occlusions in human face images make it challenging to build a robust representation model for the recognition of the query image $\y$. The sparsity is effective in sample selection for representation, and the correlation structure helps to find the relationship between the query image and training samples. Therefore, to benefit from both sparsity and correlation, we consider the structure of $\X$ as well as the sparsity of the coding coefficient $\bm{\alpha}$ by a correlation adapter denoted as $||\X\Diag(\bm{\alpha})||_{\ast}$. The main difference between the trace norm and the existing norms is the inclusion of data matrix $\X$. To guarantee the discriminative nature of training samples selected for representation, we impose the trace norm on the adapter inspired by \cite{traceLasso}. Then our linear representation model is represented as:
\begin{equation}\label{tlcORG}
\min{||\X\Diag(\bm{\alpha})|| _{\ast}} ~~ s.t. \  \y = \X\bm{\alpha},
\end{equation}
where $||\X\Diag(\bm{\alpha})|| _{\ast}$ is the correlation regularizer, denoted as $\Omega_X (\bm{\alpha})$.
Here we discuss the adaptive reformulation of our model according to the exact structure of the dictionary. In the case that the subjects are distinct from each other, the columns in the dictionary matrix $\X$ are orthogonal, $\X^T\X=\I$. Then, we get the decomposition:
\begin{equation}
\begin{split}
||\X\Diag(\bm{\alpha})||_{\ast} =   \ & \Tr[(\X\Diag(\bm{\alpha}))^T(\X\Diag(\bm{\alpha}))]^{\frac{1}{2}}\\
=\ & \Tr[(\Diag(\bm{\alpha}))^T(\Diag(\bm{\alpha}))]^{\frac{1}{2}}\\
=\ &||\alpha||_1.
\end{split}
\end{equation}
Thus, the correlation regularizer $\Omega_X (\bm{\alpha})$ is equal to the $\ell_1$-norm. 

Then, problem (\ref{tlcORG}) is the same as the sparse coding model:
\begin{equation}\label{tlc1}
\min{||\bm{\alpha}|| _{1}} ~~ s.t. \  \y = \X\bm{\alpha}.
\end{equation}
In the case that the images of different subjects look similar to $\x_1$, that is $\X = \x_1\bm{1}^T$ and $\X^T\X=\bm{11}^T$ ($\bm{1}$ is a vector of size $n$, where all the elements are one), we can get the reformulation of $\Omega_X (\bm{\alpha})$ as:

\begin{equation}\label{l2}
\begin{split}
\Omega_X (\bm{\alpha}) =  \ & || {\x}_{1}\bm{\alpha}^T || _{\ast} \\
=& || {\x}_{1} || _{2} || \bm{\alpha} || _{2} \\
=& || \bm{\alpha}||_2,
\end{split}
\end{equation}
and (\ref{tlcORG}) can be formulated as:
\begin{equation}\label{tlc2}
\min{||\bm{\alpha}|| _{2}} ~~ s.t. \  \y = \X\bm{\alpha}.
\end{equation}
We can see that (\ref{tlc2}) is essentially equal to CRC when dealing with highly correlated images.

Generally, the images in the dictionary are neither too independent of each other nor look the same. Our model can capture the correlation structure of the training images. In other words, the sparsity of the coefficient $\bm{\alpha}$ obtained by Eqn.~(\ref{tlcORG}) efficiently balances the $||\bm{\alpha}||_1$ and $||\bm{\alpha}||_2$, denoted as:
\begin{equation}\label{l2}
|| \bm{\alpha} || _2 \leq \Omega_X (\bm{\alpha}) \leq || \bm{\alpha} || _1.
\end{equation}
The above equation indicates that  $\bm{\alpha}$ obtained by $\Omega_X$ is more sparse than the one obtained by $\ell_2$, but not as sparse as the one obtained by $\ell_1$. This means that our model can benefit from both the $\ell_1$-minimization and $\ell_2$-minimization according to the correlation in the dictionary.

Problem (5) is designed for the cases where the images have no noise. However, in real-world applications, the pixels of the images may be contaminated with occlusion and corruption. If we get the prior knowledge that the noises follow the Gaussian distribution, the original objective function can be reformulated as:
\begin{equation}\label{tlc3}
\min{||\X\text{Diag}(\bm{\alpha})|| _{\ast}} ~~ s.t. \  ||\y- \X\bm{\alpha}||_2 \leq \varepsilon.
\end{equation}
If the occlusion or corruption follows the Laplacian distribution, we consider the following problem instead:
\begin{equation}\label{tlc4}
\min{||\X\text{Diag}(\bm{\alpha})|| _{\ast}} ~~ s.t. \  ||\y- \X\bm{\alpha}||_1 \leq \varepsilon.
\end{equation}
Problem (\ref{tlc4}) is more robust to occlusion, corruption and variations than problem (\ref{tlc3}). Problem (\ref{tlc4}) is equivalent to the following problem:
\begin{equation}\label{traceLasso}
\min{|| \y - \X \bm{\alpha} || _1 + \lambda || \X \text{Diag}(\bm{\alpha}) || _{\ast}},
\end{equation}
where $\lambda >0$ is the regularization parameter. We show how to solve (14) in the next subsection.

\begin{algorithm}[t]
\label{Algoopti}
\caption{Solving Problem (\ref{traceLasso}) by ADM}
\textbf{Input:} data matrix $\X$, parameter $\lambda$.   \\
\textbf{Initialize:} $\J$, $\bm{\alpha}$, $\e$, $\y_1$, $\Y_2$, $\mu$, $\rho$, $\varepsilon$ and $\mu_{\text{max}}$.\\
\textbf{while} not converge \textbf{do}
\begin{enumerate}
  \item fix the others and update $\J$ by
	\begin{equation*}
	\J=\arg\min_{\J} \frac{\lambda}{\mu}||\J||_*+\frac{1}{2}\left\| \J-(\X\text{Diag}(\bm{\alpha})-\frac{1}{\mu}{\Y}_{2})\right\| _F^2.
	\end{equation*}

  \item fix the others and update $\bm{\alpha}$ by
	\begin{equation*}
    \bm{\alpha}=\A{\X}^T\left( \frac{{\y}_1}{\mu}+\y-\e\right) +\A\text{diag}\left( {\X}^T\left( \frac{{\Y}_2}{\mu}+\J\right) \right) ,
	\end{equation*}
    where $\A=(\X^T\X+\text{Diag}(\text{diag}(\X^T\X)))^{-1}$.
  \item fix the others and update $\e$ by
	\begin{equation*}
      \e =  \arg\min_{\e} \frac{1}{\mu}||\e||_1+\frac{1}{2}\left\| \e-\left( \y-\X\bm{\alpha}+\frac{1}{\mu}{\y}_1\right) \right\| _2 .\\
    \end{equation*}

  \item update the multipliers
	\begin{equation*}
		\begin{split}
    		&{\y}_1={\y}_1+\mu(\y-\X\bm{\alpha}-\e) ,\\
    		&{\Y}_2={\Y}_2+\mu(\J-\X\Diag(\bm{\alpha})).
    	\end{split}
    \end{equation*}
  \item update the parameter $\mu$ by
	\begin{equation*}
     \mu=\min(\rho\mu,\mu_{max}).
     \end{equation*}
  \item check the convergence conditions
  \begin{equation*}
  ||\y-\X\bm{\alpha}-\e||_{\infty} \leq \varepsilon \text{  and   } ||\J-\X\Diag(\bm{\alpha})||_{\infty} \leq \varepsilon.
  \end{equation*}

\textbf{end while}
\end{enumerate}
\label{Alg_alm}
\end{algorithm}

\subsection{Optimization}
Inspired by the optimization method used in \cite{RPCA}\cite{LRR}, we adopt Alternating Direction Method (ADM) \cite{ALMlin} to solve  problem (\ref{traceLasso}). We first convert it to the following equivalent problem:
\begin{equation}\label{op1}
  \begin{split}
   \min_{\J,\bm{\alpha},\e}  \  & || \e || _1+ \lambda || \J || _{\ast}    \\
    \text{s.t.} \  & \y = \X\bm{\alpha} + \e, \\
    			   & \J = \X\Diag(\bm{\alpha}).
    \end{split}
 \end{equation}
Problem (\ref{op1}) can be solved by solving the following augmented Lagrange multiplier problem:
\begin{equation}\label{Lagfun}
\begin{split}
        L(\J,\bm{\alpha},\e) =~& ||\e||_1+\lambda||\J||_*                                   \\
        & +{\y}_1^T(\y-\X\bm{\alpha}-\e) + \Tr[{\Y}_2^T(\J-\X\Diag(\bm{\alpha}))]      \\
        & +\frac{\mu}{2}\left( ||\y-\X\bm{\alpha}-\e||_2^2+||\J-\X\Diag(\bm{\alpha})||_F^2\right) ,
    \end{split}
\end{equation}
where ${\y}_1$ and ${\Y}_2$ are Lagrange multipliers and $\mu>0$ is a parameter. This problem is unconstrained and can be minimized with respect to $\J$, $\bm{\alpha}$ and $\e$, respectively, by fixing the other variables, and then updating ${\y}_1$ and ${\Y}_2$.

Updating $\J$ when $\bm{\alpha}$ and $\e$ are fixed in Step 1 is equivalent to solving the following problem:
\begin{equation}
\begin{split}
{\J}^* = & \arg\min_{\J} L(\J,\bm{\alpha},\e) \\
 = &\arg\min_{\J} \lambda||\J||_*+\Tr({\Y}_2^T{\J})+\frac{\mu}{2}||\J-\X\Diag(\bm{\alpha})||_F^2 \\
 = & \arg\min_{\J} \frac{\lambda}{\mu}||\J||_*+\frac{1}{2}\left\| \J-[  \X\Diag(\bm{\alpha})-\frac{1}{\mu}{\Y}_2]  \right\| _F^2. \\
\end{split}
\end{equation}
The above problem has a closed form solution by the Singular Value Thresholding (SVT) operator \cite{SVT}.

Updating $\bm{\alpha}$ when fixing $\J$ and $\e$ in Step 2 is equivalent to solving the following problem:
\begin{equation}
\begin{split}
\bm{\alpha}^*= & \arg\min_{\bm{\alpha}} L(\J,\bm{\alpha},\e) \\
 = & \arg\min_{\bm{\alpha}} -{\y}_1^T{\X}\bm{\alpha}-\Tr({\Y}^T_2{\X}\Diag(\bm{\alpha}))\\
 &+\frac{\mu}{2}(\bm{\alpha}^T{\X}^T{\X}{\bm{\alpha}}-2(\y-\e)^T{\X}\bm{\alpha}
 \\&+\Tr\Diag(\bm{\alpha}){\X}^T{\X}\Diag(\bm{\alpha}))-2\Tr({\J}^T{\X}\Diag(\bm{\alpha})) \\
 = & \arg\min_{\bm{\alpha}}  \frac{\mu}{2}\bm{\alpha}^T\left({\X}^T\X+\text{Diag(diag}(X^TX))\right)\bm{\alpha}-  \\
   & \left( {\X}^T{\y}_1+\mu{\X}^T\X(\y-\e)+\text{diag}({\Y}_2^T\X+\mu {\J}^T{\X})\right) ^T\bm{\alpha}.
\end{split}
\end{equation}
The above problem can be easily solved by:
\begin{equation}
\bm{\alpha}^*=\A {\X}^T\left( \frac{1}{\mu}{\y}_1+\y-\e\right) +\A\text{diag}\left( {\X}^T\left( \frac{1}{\mu}{\Y}_2+\J\right) \right) ,
\end{equation}
where $\A=({\X}^T\X+\text{Diag}(\text{diag}({\X}^T{\X})))^{-1}$.

Updating $\e$ when fixing $\J$ and $\bm{\alpha}$ in Step 3 is equivalent to solving the following problem:
\begin{equation}
\begin{split}
{\e}^*= & \arg\min_{\e} L(\J,\bm{\alpha},\e) \\
 = & \arg\min_{\e} ||\e||_1-{\y}_1^T\e+\frac{\mu}{2}||\y-\X\bm{\alpha}-\e||_F^2 \\
 =  & \arg\min_{\e} \frac{1}{\mu}||\e||_1+\frac{1}{2}\left\| \e-\left( \y-\X\bm{\alpha}+\frac{1}{\mu}{\y}_1\right) \right\| _2. \\
\end{split}
\end{equation}
The solution to the above problem can be obtained by the soft-thresholding (shrinkage) operator \cite{hale2008fixed}.
The whole algorithm for solving problem (\ref{traceLasso}) is outlined in Algorithm \ref{Alg_alm}.

\subsection{Adaptive Sparse Representation based Classification}

\begin{algorithm}[t]
\caption{Adaptive Sparse Representation based Classification}
\textbf{Input:} data matrix $\X$, the query image $\y$.\\
\textbf{Output:} the identity of the query image $\y$.
\begin{enumerate}
\item Normalize each column of $\X$ to have unit $\ell_2$-norm.
\item Code $\y$ over $\X$ by solving
\begin{equation*}\label{qiujie}
\hat{\bm{\alpha}} = \arg\min || \y - \X \bm{\alpha} || _1 + \lambda || \X \text{Diag}(\bm{\alpha}) || _{\ast}
\end{equation*}

\item Compute the residuals
\begin{equation*}
{\rr}_i = || \y-{\X}_i\hat{\bm{\alpha}}_i|| _{2},
\end{equation*}
where $\bm{\alpha_i}$ is the coding coefficient vector associated with  ${\X}_i$.
\item Predict the identity of $\y$ by
\begin{equation*}\label{error}
\text{identity}(\y) = \arg\min \{{\rr}_i\}.
\end{equation*}
\end{enumerate}

\label{TLC}
\end{algorithm}

Based on the robust model for coding of query images defined in problem (\ref{traceLasso}), we present our Adaptive Sparse Representation based Classification (ASRC) method for face recognition. The scheme of the algorithm is described in Algorithm 2. First, we normalize each column of the dictionary matrix $\X$ in Step 1. Given the query image $\y$, we code it over the whole dictionary with correlation regularizer $||\X\text{Diag}(\bm{\alpha})||_{\ast}$ in Step 2. The optimal coefficient $\hat{\bm{\alpha}}$ can be solved by ADM described in Algorithm 1. As the correlation regularizer is adaptive to the structure of the dictionary, the sparsity of the coefficient $\hat{\bm{\alpha}}$ balances the coefficients obtained by the $\ell_1$-minimization and $\ell_2$-minimization. Thus, the nonzero entities of $\hat{\bm{\alpha}}$ will focus on discriminative training samples. In Step 3, we calculate the residuals of different subjects. The subject to which $\y$ belongs will give a better representation, leading to smaller representation error. Then, the query image $\y$ will be assigned to the class with the least residual in Step 4. It can be seen that the time complexity of the algorithm is $O(n^3)$.

Based on the above analysis, we can observe that our algorithm has some attractive properties:
\begin{itemize}
\item First, our algorithm can obtain an accurate representation of the query image according to the structure of the dictionary, which is critical for representation based methods in face recognition.

\item Second, ASRC can benefit from both the discriminative nature of $\ell_1$-norm and the collaborative representation of $\ell_2$-norm, which will guarantee a good recognition performance in most cases.

\item Third, due to the adaptive correlation regularizer, the information loss, such as misalignment, pixel corruption or insufficient training samples, can be compensated by the correlation of training samples. Even if the dictionary has limited training samples per class, we can still obtain an accurate representation of the query image compared with sate-of-the-art methods, such as SRC and CRC.
\end{itemize}

\section{Experiments}
\label{sec4}
In this section, we first investigate the recognition performance and robustness of our proposed ASRC method for face recognition. Later, we design experiments to test the generalization ability of ASRC on general pattern recognition problems. Two sets of databases are used in the experiments, one from the real-world face image databases, including Yale \cite{belhumeur1997eigenfaces}, ORL \cite{samaria1994parameterisation} and AR \cite{martinez1998ar}, and the other sampled from the UCI repository \cite{Bache+Lichman:2013}. Table 1 summarizes these data sets in terms of the numbers of classes, dimensions and samples. Some exemplar face images are shown in Figure 2. For each face image database, we choose $t$ images of each subject for training and the rest for test. On the UCI data sets, we adopt 10-fold cross validation.

In the following subsections, we test the recognition performance and robustness of our algorithm by comparing it with some methods like SVM (linear kernel)~\cite{CC01a}, SRC, NN, NFS, CRC, and LSRC (Locality-constrained Sparse Representation based Classifier) \cite{wang2010locality}. We use $t$-test to test the statistic significance of the results. The significance level is set to be 0.05. We adopt PCA for feature extraction on the face image data sets.

\subsection{Face Recognition Without Occlusion}

\begin{table}[!t]
\centering
\caption{Description of 18 data sets}
\label{Tab UCI}
\centering
\begin{tabular}{|c c c c|}
\hline
\multicolumn{4}{|c|} {Face Image Data Sets}  \\
Data Set &$\#$class &$\#$dim. &$\#$instance \\ \hline
Yale &15 &1024 &165 \\ \hline
ORL &40 &1024 &400 \\ \hline
AR &100 &3168 &1400 \\ \hline  \hline
\multicolumn{4}{|c|} {UCI Data Sets}  \\
Data Set &$\#$class &$\#$dim. &$\#$instance \\ \hline
UCI Data Sets & & &\\ \hline
Diabetes & 2 & 9 & 768\\ \hline
Breast & 2 & 10 & 277 \\ \hline
Breast\_gy & 2 & 10 & 277 \\ \hline
Heart & 2 & 13 & 270 \\ \hline
Hearts & 2 & 13 & 270\\ \hline
Cleve & 2 & 14 & 296 \\ \hline
Vote & 2 & 17 & 435 \\ \hline
German & 2 & 25 & 1000 \\ \hline
Ionosphere & 2 & 35 & 351 \\ \hline
Spectf & 2 & 44 & 267 \\ \hline
Wdbc & 2 & 31 & 267 \\ \hline
Air & 3 & 65 & 359 \\ \hline
X8D5K & 5 & 9 & 1000 \\ \hline
Glass & 6 & 10 & 214 \\ \hline
Glass\_gy & 6 & 10 & 214 \\ \hline
\end{tabular}
\end{table}

\begin{figure*}[!t]
\centering
\includegraphics[width=1\textwidth]{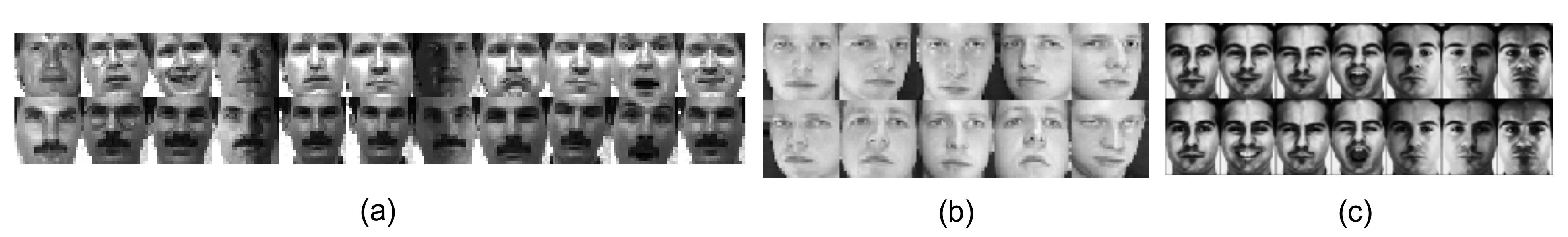}
\caption{Face image samples. (a) Two subjects in the Yale database. (b) One subject in the ORL database. (c) One subject in the AR database.}

\end{figure*}

a) Results on Yale Database

The Yale database contains 165 images of 15 individuals with various facial expressions or illuminations, as shown in Figure 2a. The images are taken under different emotions of the subject, such as sad, happy and surprised. They are cropped to 32$\times$32 pixels and converted to gray scale. In the experiment, we choose a random subset with $t$ (= 4, 5, 6, 7) images per subject to form the training set and take the rest for test. We select different feature space dimensions for various $t$ values. For each given $t$, we repeat the experiment over 10 random splits of the data set and record the average accuracy in Figure 3. We also provide the maximal accuracy and the standard deviation of each algorithm on different $t$ values in Table \ref{YALET}.
\begin{figure*}[!t]
   \begin{subfigure}[b]{0.240\textwidth}
		\centering \label{yale}
        \includegraphics[width=\textwidth]{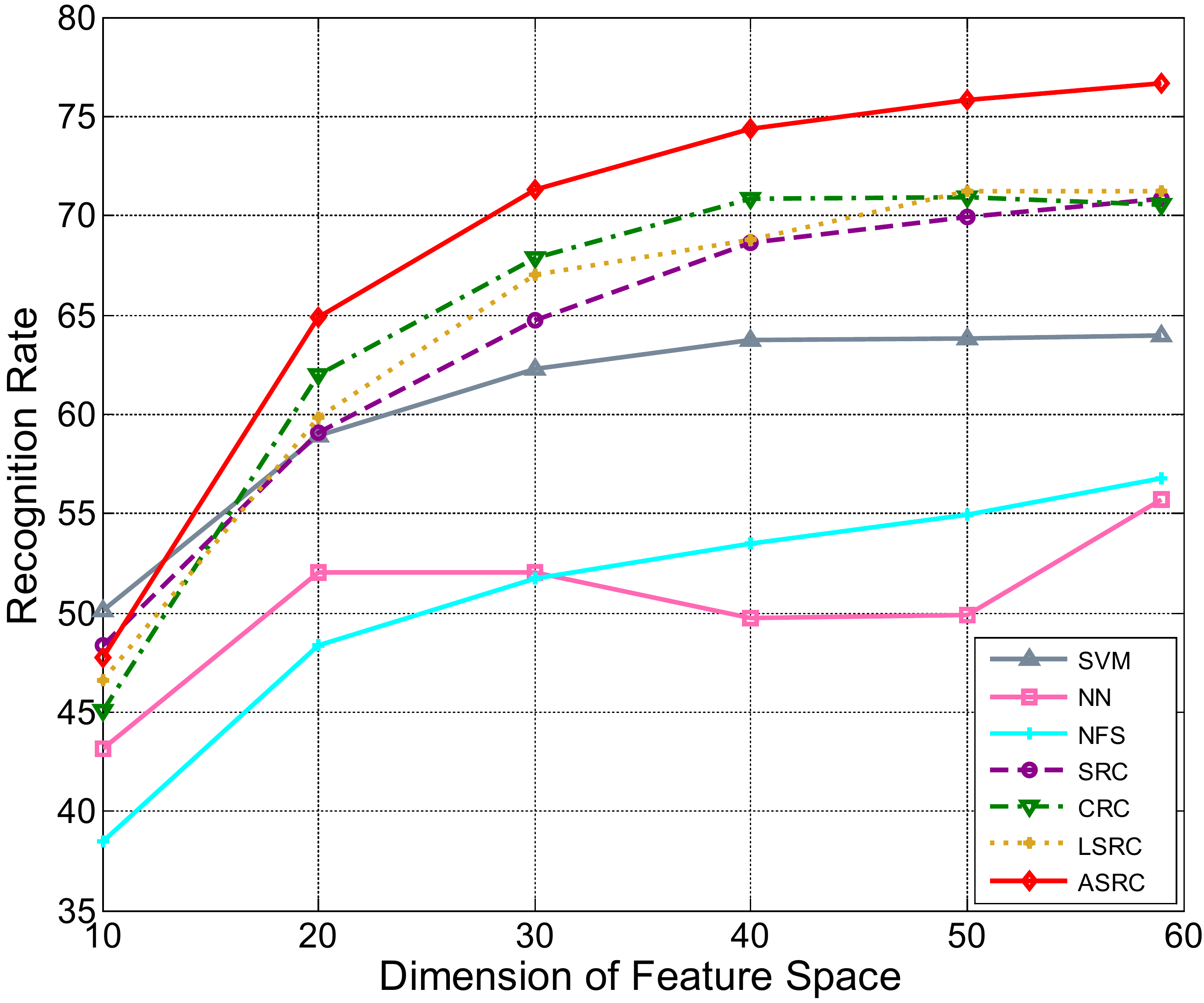}
        \caption{  $t$ = 4}
    \end{subfigure}	
     \begin{subfigure}[b]{0.24\textwidth}
		\centering  
        \includegraphics[width=\textwidth]{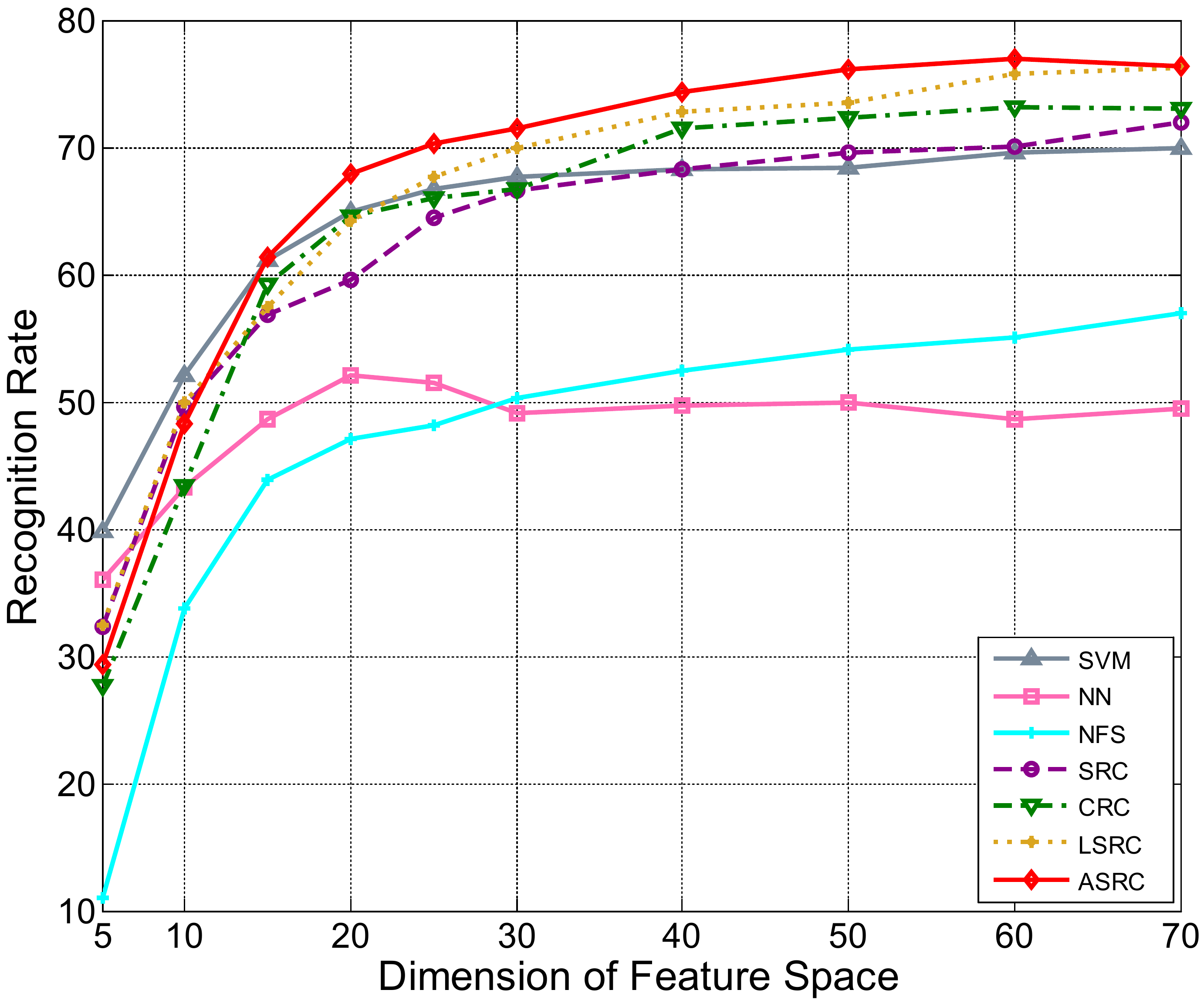}
        \caption{  $t$ = 5}
    \end{subfigure}
    \begin{subfigure}[b]{0.24\textwidth}
		\centering
		\includegraphics[width=\textwidth]{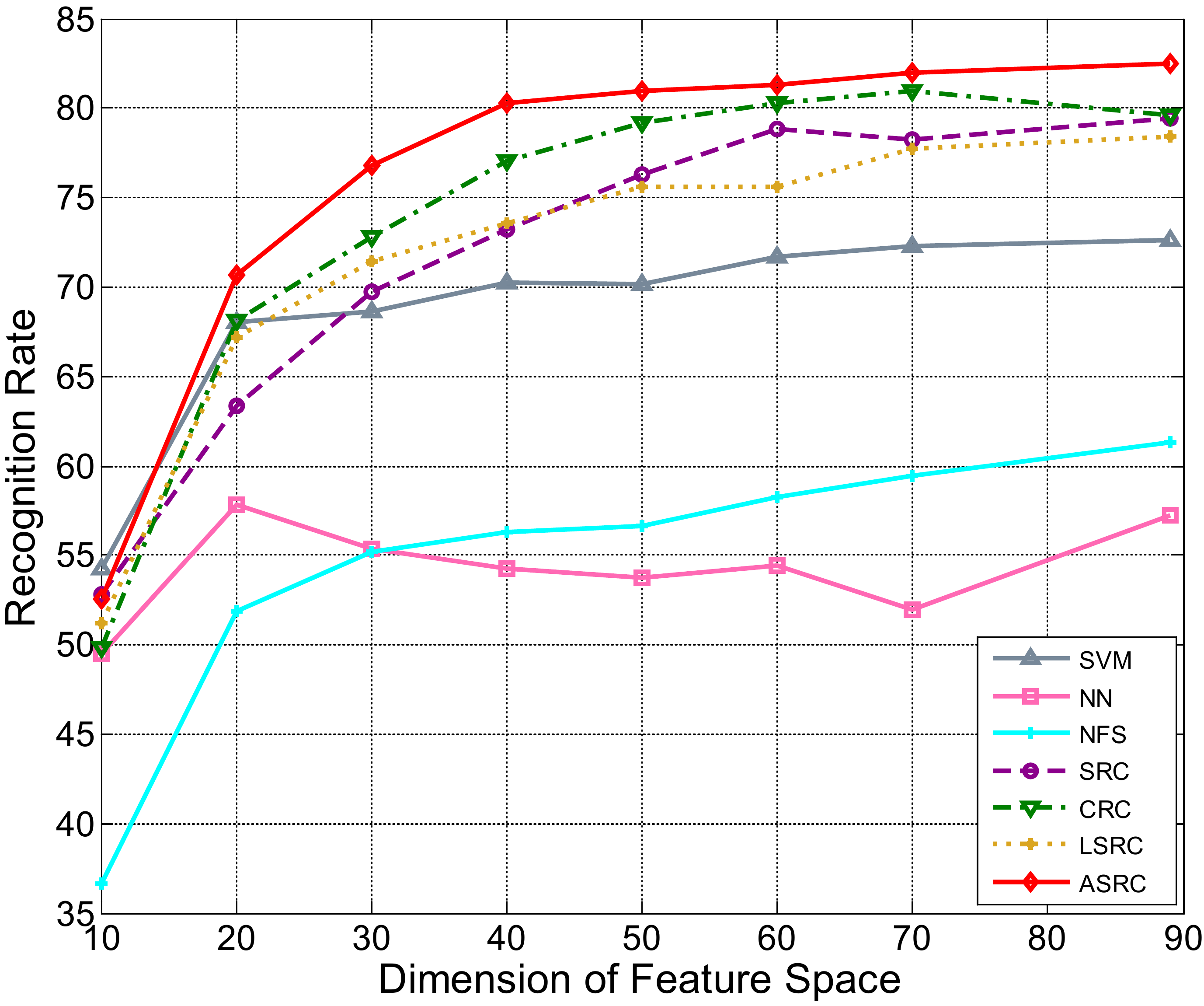}
		\caption{  $t$ = 6}
    \end{subfigure}
	\begin{subfigure}[b]{0.24\textwidth}
		\centering
        \includegraphics[width=\textwidth]{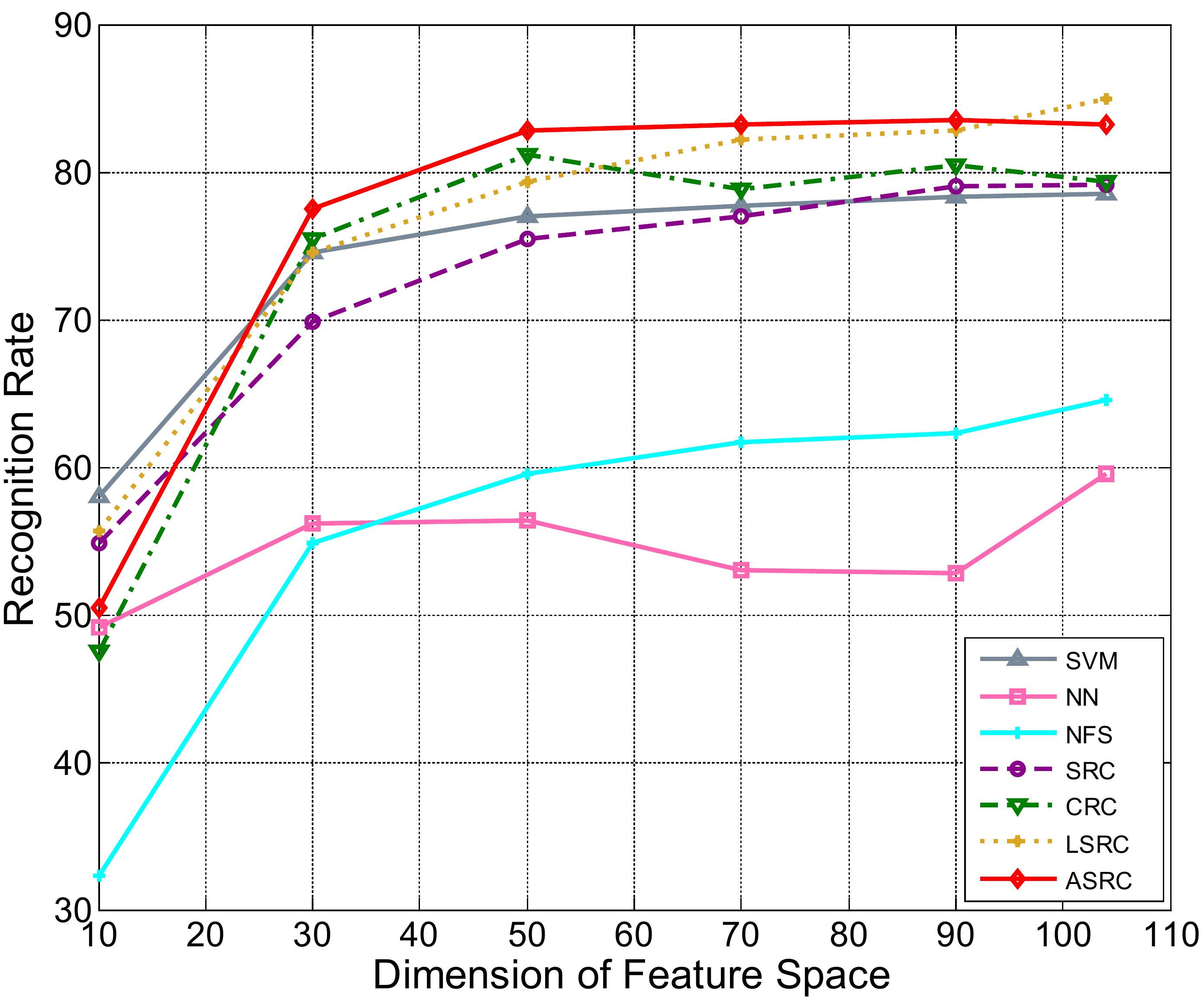}
        \caption{  $t$ = 7}
    \end{subfigure}
    \caption{Comparison recognition rates based on $t$ images of each subject for training on the Yale database.}
	\label{fig_video}
	\vspace{-1em}
\end{figure*}

\begin{table*}[!t]
\centering
\caption{The maximal average accuracy and the standard deviation of different algorithms on the Yale database vs. the dimension of the feature space when the maximal accuracy is obtained.}
\label{YALET}
\centering
\begin{tabular}{|c|c|c|c|c|}
\hline
Algorithms & $t$ = 4 & $t$ = 5 &$t$ = 6 & $t$ = 7\\ \hline\hline
SVM 	& 64.00$\pm$2.57 (59) 	& 69.89$\pm$5.48 (70) 	& 72.67$\pm$3.28 (89) 	& 78.50$\pm$6.73 (104) \\ 
NN 		& 55.71$\pm$4.65 (59)	& 52.11$\pm$4.30 (20)	& 57.87$\pm$4.92 (20)	& 59.50$\pm$3.69 (104) \\ 
NFS 	& 56.76$\pm$5.30 (59)	& 57.00$\pm$4.74 (70)	& 61.33$\pm$5.96 (89)    & 64.50$\pm$5.21 (104)   \\ 
SRC 	& 70.86$\pm$4.56 (59)	& 72.00$\pm$4.02 (70)	& 79.47$\pm$3.68 (89)	& 79.17$\pm$3.17 (104)  \\ 
CRC 	& 70.95$\pm$4.67 (50)	& 73.11$\pm$4.79 (60)	& 80.93$\pm$3.93 (70)	& 81.17$\pm$3.93 (50) \\ 
LSRC    & 71.24$\pm$2.49 (50)	& 76.22$\pm$3.93 (70)	& 78.40$\pm$3.86 (89)	& 85.00$\pm$5.56 (104) \\ 
ASRC    & \textbf{76.67}$\pm$4.71 (59) & \textbf{77.00}$\pm$4.26 (60) & \textbf{82.53}$\pm$2.67 (89)	& \textbf{83.17}$\pm$4.89 (104) \\ \hline
\end{tabular}
\end{table*}

It can be seen from Figure 3 that ASRC outperforms the other methods at all levels. The improvement gain of ASRC appears to be more significant when the number of training samples is limited ($t$ = 4, 5). The reason is that ASRC considers both the correlation and sparsity. Even with insufficient training samples, the variations in the query image can be captured by selecting sufficient correlated training samples. Meanwhile, compared with CRC, ASRC only chooses the most discriminative samples for representation which will lead to higher recognition rates. From Figure 3c and Figure 3d, when there are relatively enough training samples, we can see that ASRC, SRC and CRC all perform well. SRC, LSRC and CRC have similarly good results. LSRC outperforms SRC in most cases and obtains better results than both SRC and CRC when $t=5$. The reason is that LSRC considers the local information of the dictionary. Though LSRC obtains better performance, LSRC, SRC and CRC are all inferior to our method. This is because our algorithm ASRC considers the exact structure information of the dictionary and makes the model adaptive to the structure. Thus, with complementary information, our algorithm obtains the best results. Compared with NN and NFS, the recognition rates of our method are at least 10\% higher. SVM obtains much better results compared with NN and NFS, but is still inferior to our algorithm. To sum up, the experimental results show that the adaptive balance between sparsity and correlation does contribute to face recognition significantly.

b) Results on ORL Database

The ORL database contains face images of 40 distinct subjects captured in different time with variations in illumination, facial expression and details (glasses), as shown in Figure~2b. There are no restrictions imposed on the expression but the side movement or tilt is controlled within 20 degrees. For each subject, we select $t$ (= 2, 3, 4, 5) images for training and use the rest for test. The average accuracy rates versus selected features are recorded over 10 random splits, summarized in Figure 4. We also report the maximal accuracy and the standard deviation of each algorithm on the ORL database in Table \ref{TabORL}.
\begin{figure*}[!t]\label{orl}
   \begin{subfigure}[b]{0.23\textwidth}
		\centering
        \includegraphics[width=\textwidth]{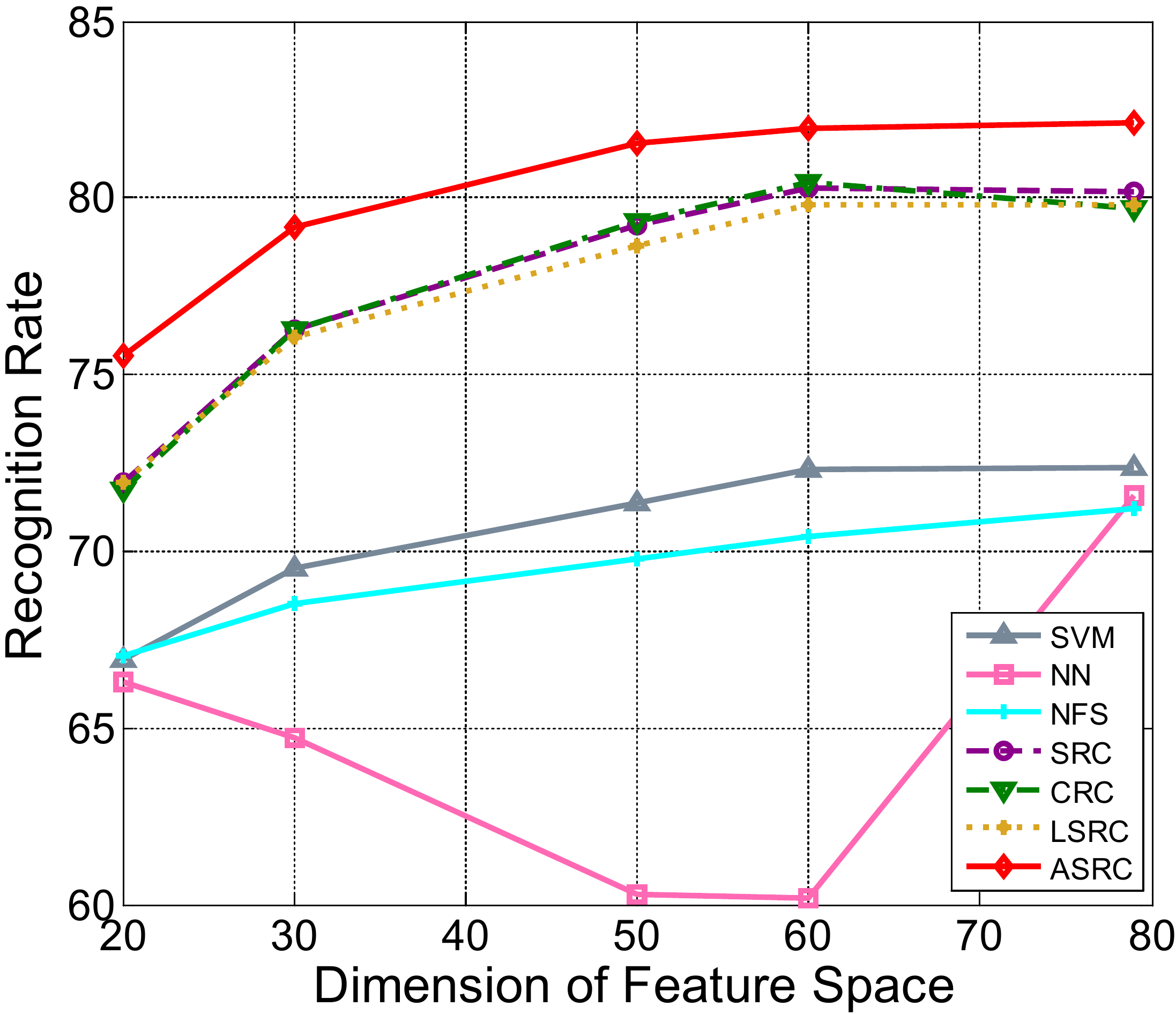}
        \caption{  $t$ = 2}
    \end{subfigure}	
     \begin{subfigure}[b]{0.25\textwidth}
		\centering  
        \includegraphics[width=\textwidth]{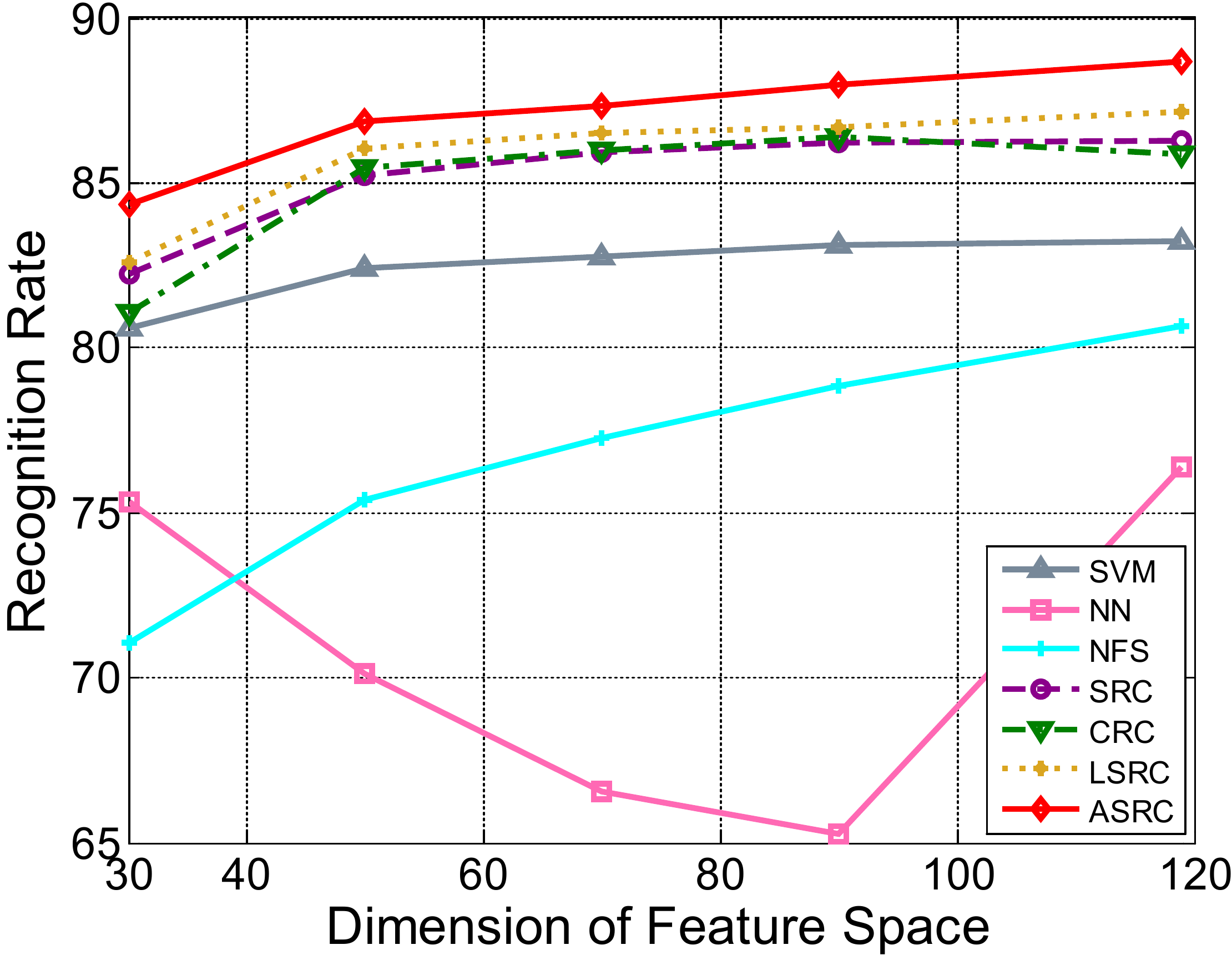}
        \caption{  $t$ = 3}
    \end{subfigure}
    \begin{subfigure}[b]{0.25\textwidth}
		\centering
		\includegraphics[width=\textwidth]{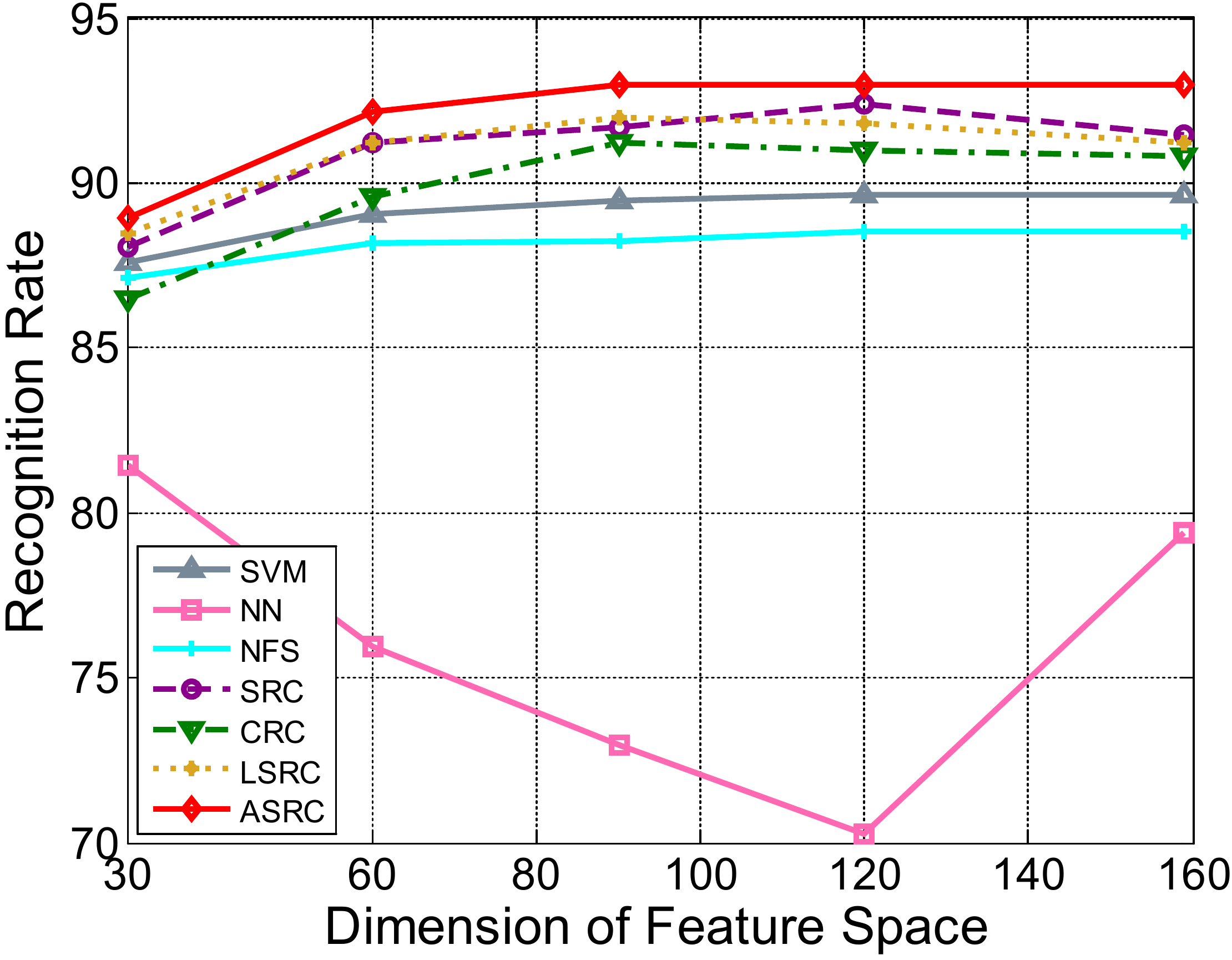}
		\caption{  $t$ = 4}
    \end{subfigure}
	\begin{subfigure}[b]{0.255\textwidth}
		\centering
        \includegraphics[width=\textwidth]{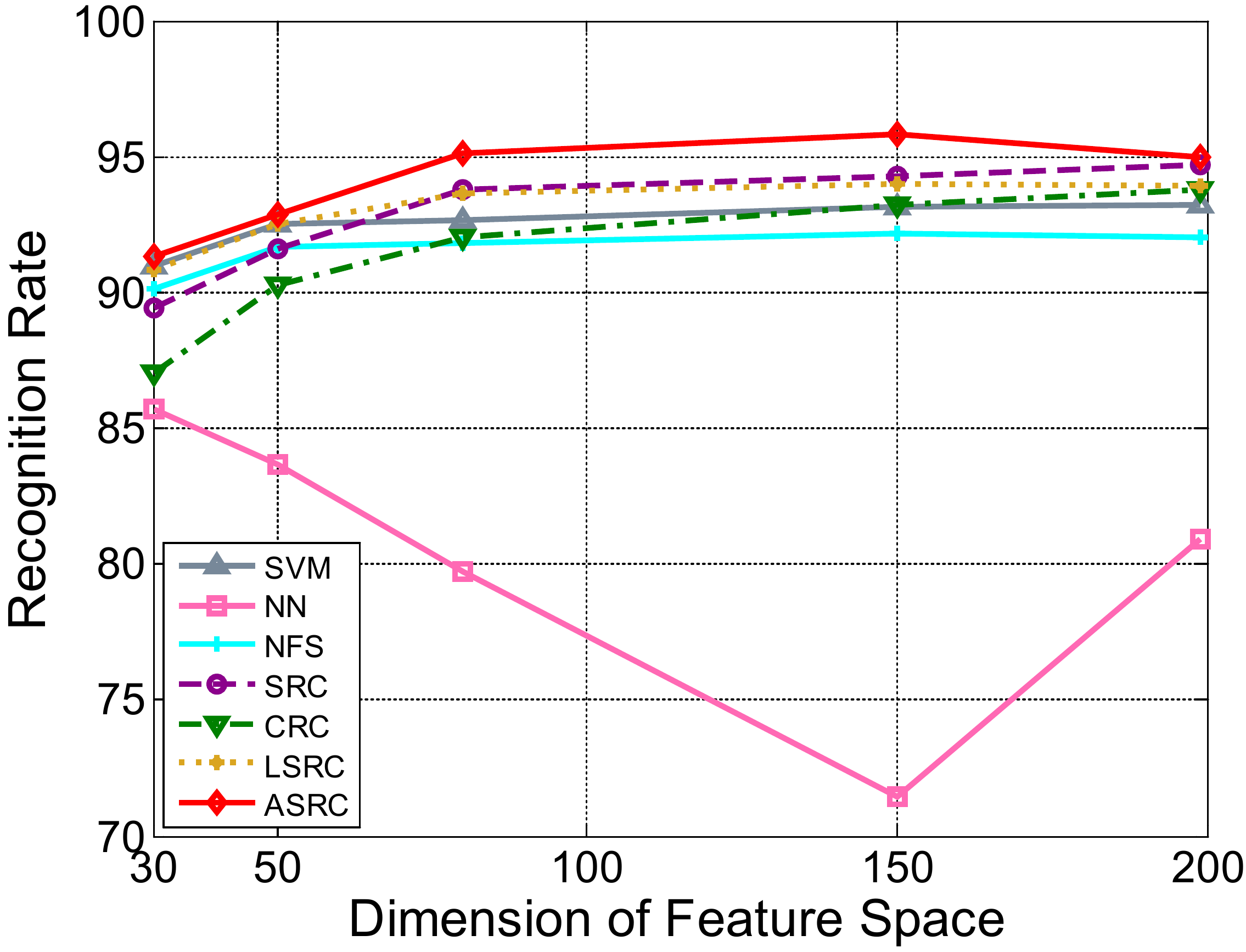}
        \caption{  $t$ = 5}
    \end{subfigure}
    \caption{Comparison recognition rates based on $t$ images of each subject for training on the ORL database.}

	\vspace{-1em}
\end{figure*}

\begin{table*}[!t]
\centering
\caption{The maximal average accuracy and the standard deviation of different algorithms on the ORL database vs. the dimension of the feature space when the maximal accuracy is obtained.}
\label{TabORL}
\centering
\begin{tabular}{|c|c|c|c|c|}
\hline
Algorithms & $t$ = 2 & $t$ = 3 & $t$ = 4 &$t$ = 5\\ \hline\hline
SVM 		& 72.38$\pm$4.00 (79) 		& 83.25$\pm$1.94 (119) 	& 89.67$\pm$1.90 (120)	 & 93.20$\pm$1.60 (199) \\ 
NN 			& 71.59$\pm$3.23 (79) 		& 76.36$\pm$2.37 (119) 	& 81.46$\pm$1.89 (30)	 & 85.70$\pm$2.37 (30) \\ 
NFS 		& 71.19$\pm$3.48 (79)		& 80.64$\pm$1.70 (119)	& 88.54$\pm$2.03 (120)	 & 92.20$\pm$1.72 (120)  \\ 
SRC 		& 80.28$\pm$2.52 (60)		& 86.29$\pm$1.58 (119)	& 92.37$\pm$0.88 (120) 	& 94.70$\pm$1.44 (199)  \\ 
CRC 		& 80.44$\pm$2.41 (60)		& 86.39$\pm$2.07 (90)	& 91.21$\pm$1.72 (90)	& 93.75$\pm$2.12 (199) \\ 
LSRC		& 79.81$\pm$2.46 (60)		& 87.14$\pm$1.87 (119)	 & 92.00$\pm$1.22 (90) & 94.00$\pm$2.12 (150) \\ 
ASRC		& \textbf{81.69}$\pm$2.86 (79) 	   & \textbf{88.68}$\pm$2.03 (119)   & \textbf{93.00}$\pm$0.87 (159)       & \textbf{95.85}$\pm$2.12 (150) \\ \hline
\end{tabular}
\end{table*}

From Figure 4 and Table \ref{TabORL}, we can see that ASRC obtains the best recognition rates at all levels. The performances of all the methods improve as the training samples increase, and our method always remains the best. When there are insufficient training samples ($t$ = 2), SRC, CRC and LSRC have similar performances. The reason is that even though CRC considers the correlation, it does not perform sample selection for representation which will disturb the final recognition results. LSRC only considers the local information which is limited with insufficient training samples. Our method takes advantage of the correlation of the query image and the training samples, thus it can obtain relatively more information. Compared with these methods, our method balances the sparsity and the correlation. Thus, ASRC can obtain stable and relatively better recognition rates. From Table \ref{TabORL}, we can see that the algorithms may not obtain the maximum average accuracy when the dimension is the largest. For instance, in the case of $t=5$, NN obtains the maximum average accuracy when the dimension of the feature space is 30, SVM, NFS and SRC when 120, and CRC and LSRC when 90. Our algorithm always obtains the best result when the dimension reaches the largest, and the recognition rates of our accuracy are also higher than the others.

c) Results on AR Database

The AR database consists of over 4000 frontal images of 126 subjects. In this experiment, a subset (with only illumination and expression changes) that contains 50 male subjects and 50 female subjects is chosen from the AR database, as shown in Figure 2c. The images are cropped to 165$\times$120 pixels. For each subject, we choose $t$ (= 2, 7) images for training and take the rest for test. The experimental results are shown in Figure 5 and Table \ref{TabAR}.
\begin{figure*}[!t]\label{ar}
   \begin{subfigure}[b]{0.48\textwidth}
		\centering
        \includegraphics[width=\textwidth]{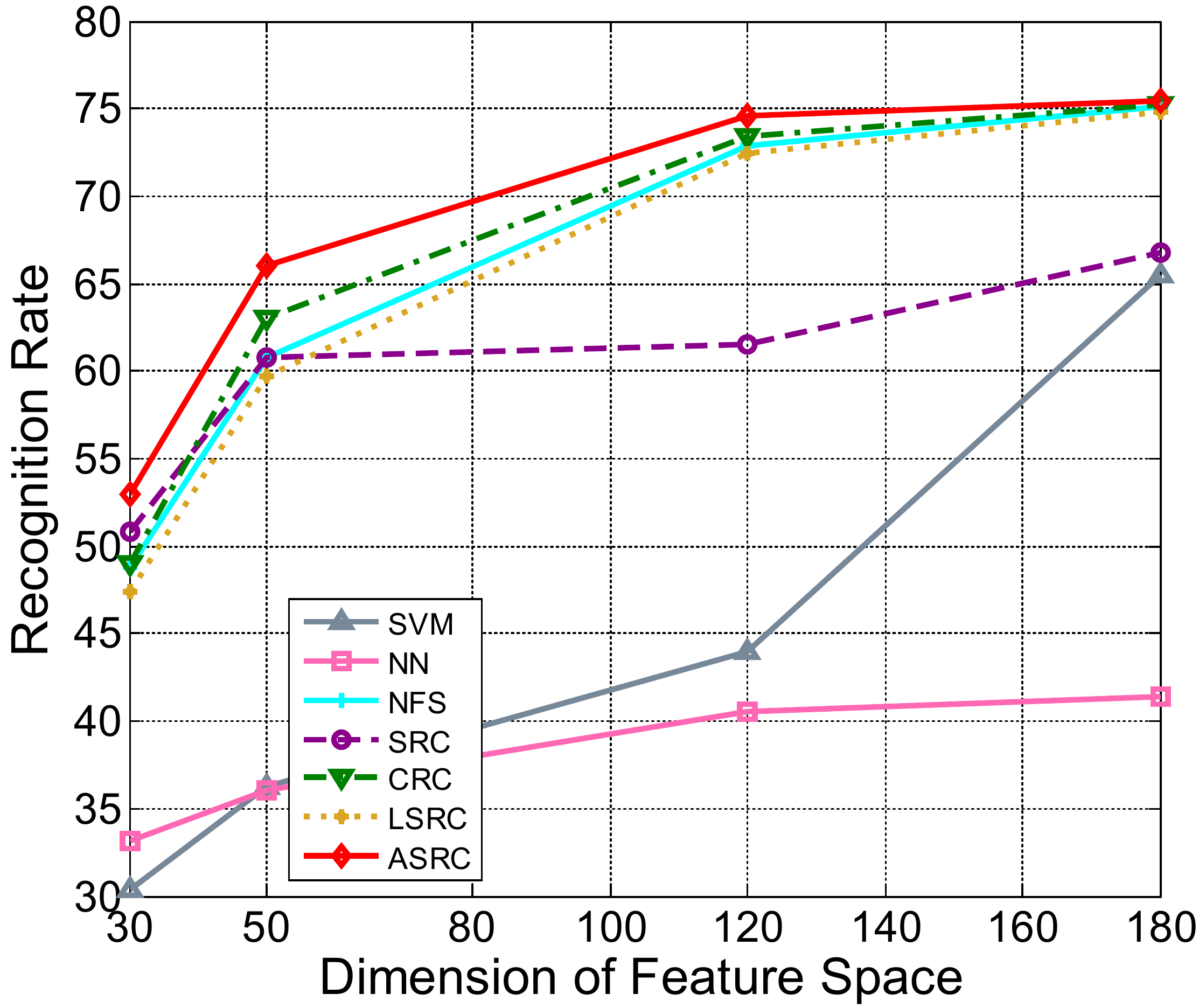}
        \caption{  $t$ = 2}
    \end{subfigure}	
     \begin{subfigure}[b]{0.5\textwidth}
		\centering  
        \includegraphics[width=\textwidth]{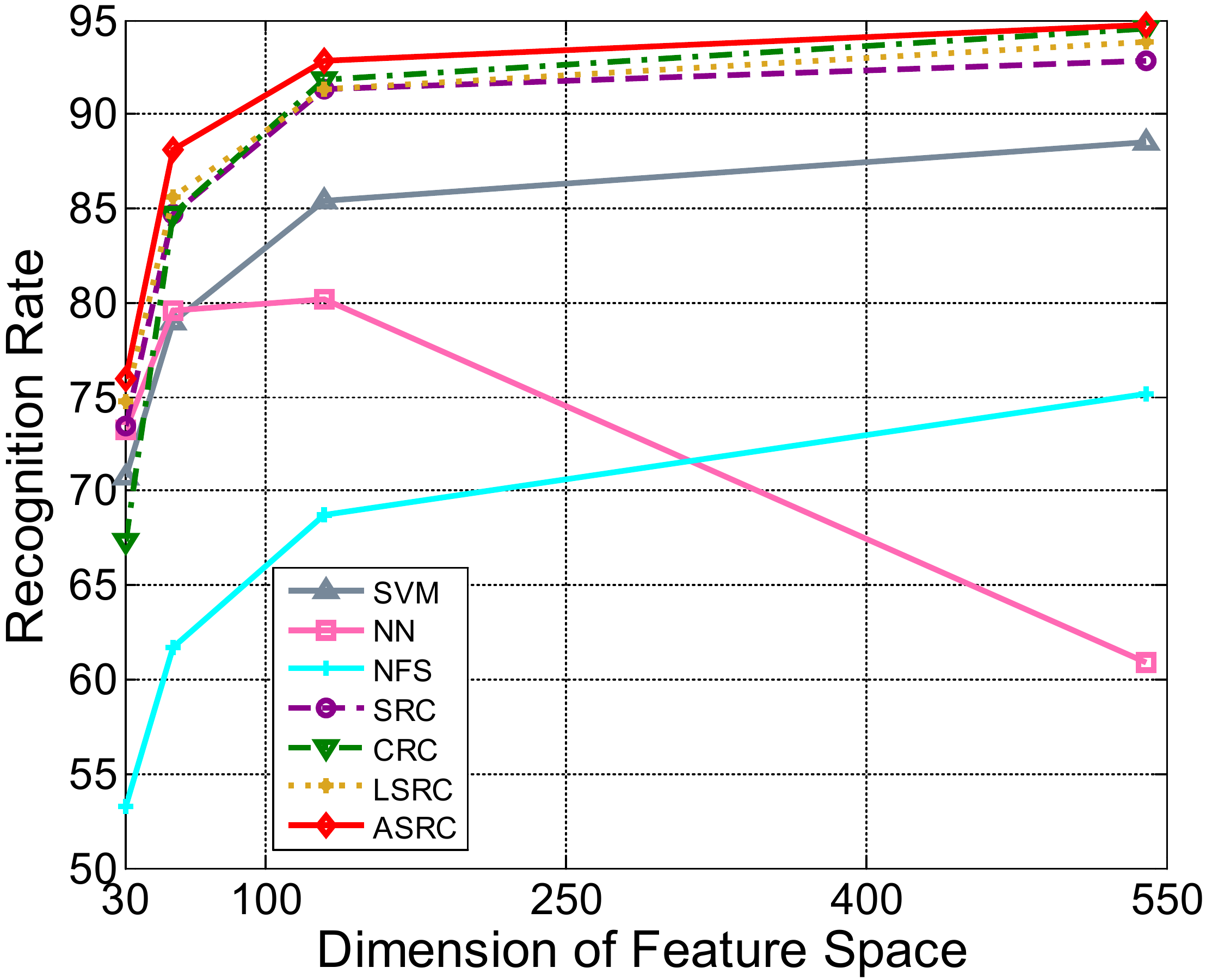}
        \caption{  $t$ = 7}
    \end{subfigure}
    \caption{Comparison recognition rates based on $t$ images of each subject for training on the AR database.}

	\vspace{-1em}
\end{figure*}

It can be seen from Figure 5 and Table \ref{TabAR} that our method still obtains the best results in all situations. The performance of NFS is very unstable. More specifically, in the case where $t=2$, NFS outperforms SRC, LSRC, SVM and NN. In another case where $t=7$, NFS is inferior to all the other algorithms when the dimension is less than 250. This is because NFS only depends on the representation  of subspaces which is easily affected by the disturbance of the training samples. CRC and LSRC obtain relatively much stable results in both cases compared with SRC, as more information is considered. Most of the algorithms obtain the maximum average accuracy when the dimension of the feature space reaches the largest. Generally speaking, with sufficient features, the algorithms can obtain much better results. More specifically, the maximum average accuracy of ASRC when $t=2$ is $75.50 \%$, and it reaches $94.71\%$ when $t=7$.

Based on the results shown in Figures 3, 4 and 5 and the Tables \ref{YALET}, \ref{TabORL} and \ref{TabAR}, we can draw the following conclusions:
\begin{itemize}
\item For all the three databases, the best performance of ASRC consistently exceeds those of the competing methods. More specifically, when $t = 4$, the best recognition rate for ASRC on the Yale database is 76.67 $\%$, compared to 55.57 $\%$ for NN, 56.76 $\%$ for NFS, 70.86 $\%$ for SRC, 70.95 $\%$ for CRC and 71.24 $\%$ for LSRC; when $t = 5$, the best rate for ASRC on the ORL database is 95.58 $\%$, compared to 85.7 $\%$ for NN, 92.2 $\%$ for NFS, 94.7 $\%$ for SRC, 93.75 for CRC and 94.00 $\%$ for LSRC; when $t = 2$, the best rate for ASRC on the AR database is 75.5 $\%$, 66.83 $\%$ for NN, 41.42 $\%$ for NFS, 75.25 $\%$ for SRC, 75.25 $\%$ for CRC and 74.83 $\%$ for LSRC.

\item Generally, SRC, CRC and LSRC have stable good performances in most cases. Among them, when the training samples are insufficient (Figure 3a and Figure 4a), CRC tends to have better performance due to relatively sufficient information as more samples will be selected for presentation. LSRC outperforms the other two algorithms as shown in Figure 3b and Figure 4b, due to its consideration of both sparsity and locality. However, the local information is not sufficient. Our method ASRC, which considers the exact structure and correlation information of the dictionary, yields the best recognition results in most cases. It also outperforms the classical SVM algorithm and other NFCs. Thus the experimental results demonstrate that our algorithm outperforms state-of-the-art face recognition methods. 
\end{itemize}

\subsection{Face Recognition Despite Random Pixel Corruption}
In this subsection, we test the robustness of ASRC on the three face data sets. The images in Yale and ORL are resized to 16$\times$16 pixels, and the images in AR are resized to 66$\times$48 pixels. For the Yale database, we choose 6 images per subject for training and use the rest for test. For ORL and AR databases, we randomly select half of the images for training and use the other half for test.

\begin{table}[!t]
\centering
\caption{The maximal average accuracy and the standard deviation of different algorithms on the AR database vs. the dimension of the feature space when the maximal accuracy is obtained.}
\label{TabAR}
\centering
\begin{tabular}{|c|c|c|}
\hline
Algorithms & $t$ = 2  & $t$ = 7\\ \hline\hline
SVM 		& 65.50 (180)  		    & 88.57 (540)	 	\\ 
NN 			& 41.42 (180)	        & 80.14 (130)				 \\ 
NFS 		& 75.17 (180)			& 75.14 (540)				\\ 
SRC 		& 66.83 (180)			& 92.86 (540)					 \\ 
CRC			& 75.25 (180)			& 94.57 (540)				\\ 
LSRC		& 74.83 (180)			& 93.86 (540)		\\ 
ASRC		& \textbf{75.50} (180)		    & \textbf{94.71} (540)  \\ \hline
\end{tabular}
\end{table}
We corrupt a certain percentage of the randomly chosen pixels in each of the test images, replacing their values with independent distributed samples from a uniform distribution. The corrupted pixels are randomly chosen for each test image. The percentage of corrupted pixels varies from 10 $\%$ to 80 $\%$. Figure 6 shows several example test images of three subjects.

We compare our method with four popular techniques to test its robustness. 
Figure 6 plots the recognition performance of ASRC and its 6 competitors over various levels of corruption. From Figure 6a which depicts the comparison results on the Yale database, we can see that the proposed algorithm dramatically outperforms others. For 0 $\%$ up to 20 $\%$ occlusion, our algorithm recognizes the subjects with recognition rate of over 80 $\%$. For 30 $\%$ to 40~$\%$ occlusion, the recognition rates of our algorithm are over 10 $\%$ higher than other competitors. For roughly 50 $\%$ and 60 $\%$ occlusion, ASRC still obtains the best recognition rates, at least 3 $\%$ higher than other competitors. On the ORL database (Figure 6b), our algorithm achieves the recognition rate of over 85 $\%$ when the occlusion increases from 0 $\%$ to 30 $\%$. At 40 $\%$ corruption, none of the compared algorithms achieves higher than 70 $\%$ recognition rate, while the proposed algorithm achieves 78~$\%$. Even at 50~$\%$ occlusion, the recognition rate is still over 60 $\%$. Figure 6c plots the recognition performance of ASRC and its competitors on the AR database. From 0 $\%$ up to 40~$\%$ occlusion, ASRC and SRC correctly classify the subjects with recognition rate of around 80 $\%$, much better than the other methods. In some cases on the Yale and AR databases, especially when the occlusion reaches 60 $\%$, some algorithms (LSRC, SRC) reach slightly higher recognition rates than ASRC. However, the difference is not statistically significant.

\begin{figure*}[!t]\label{noise}
   \begin{subfigure}[b]{0.33\textwidth}
		\centering
        \includegraphics[width=\textwidth]{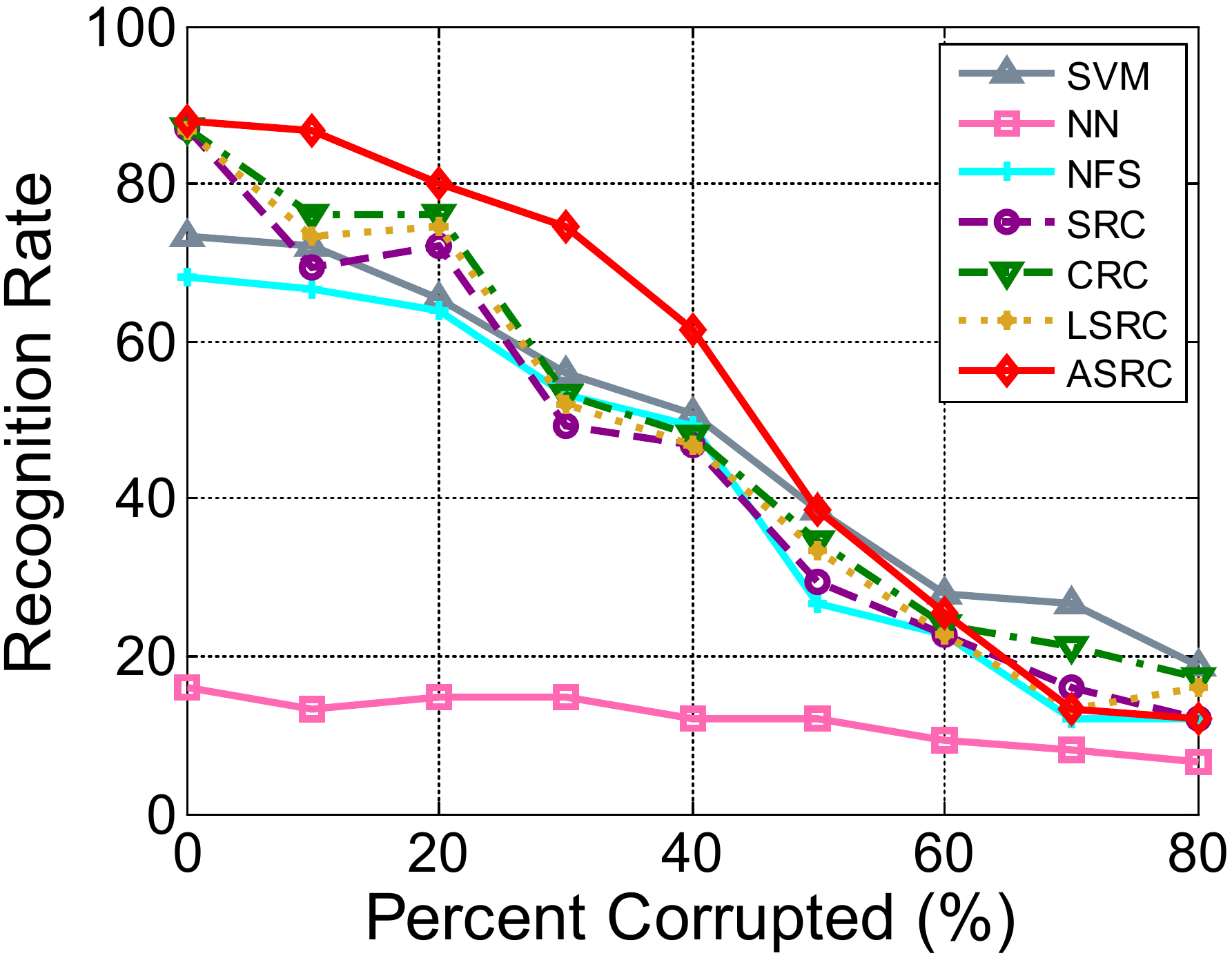}
        \caption{ Yale}
    \end{subfigure}	
     \begin{subfigure}[b]{0.33\textwidth}
		\centering  
        \includegraphics[width=\textwidth]{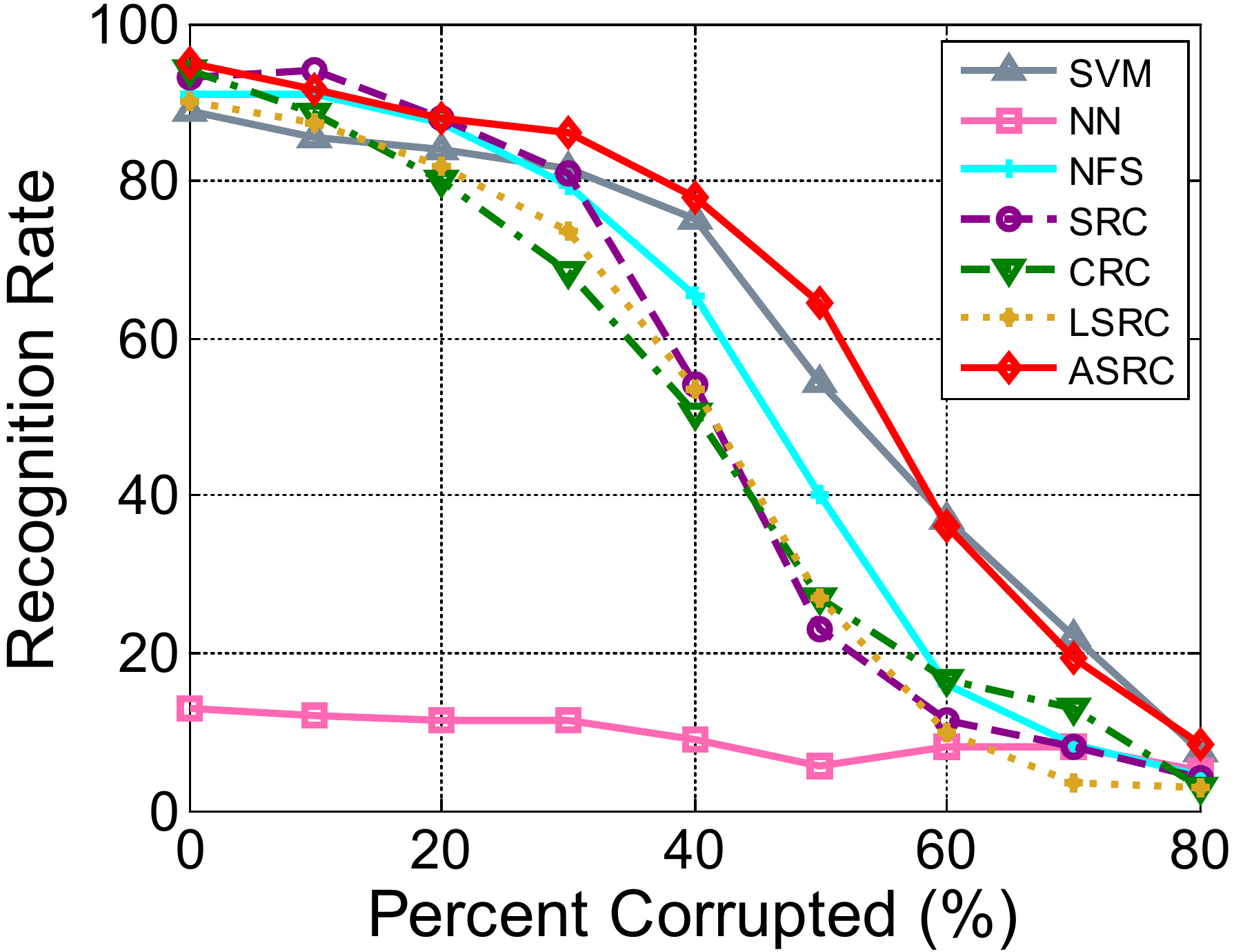}
        \caption{ ORL}
    \end{subfigure}
     \begin{subfigure}[b]{0.33\textwidth}
    		\centering  
            \includegraphics[width=\textwidth]{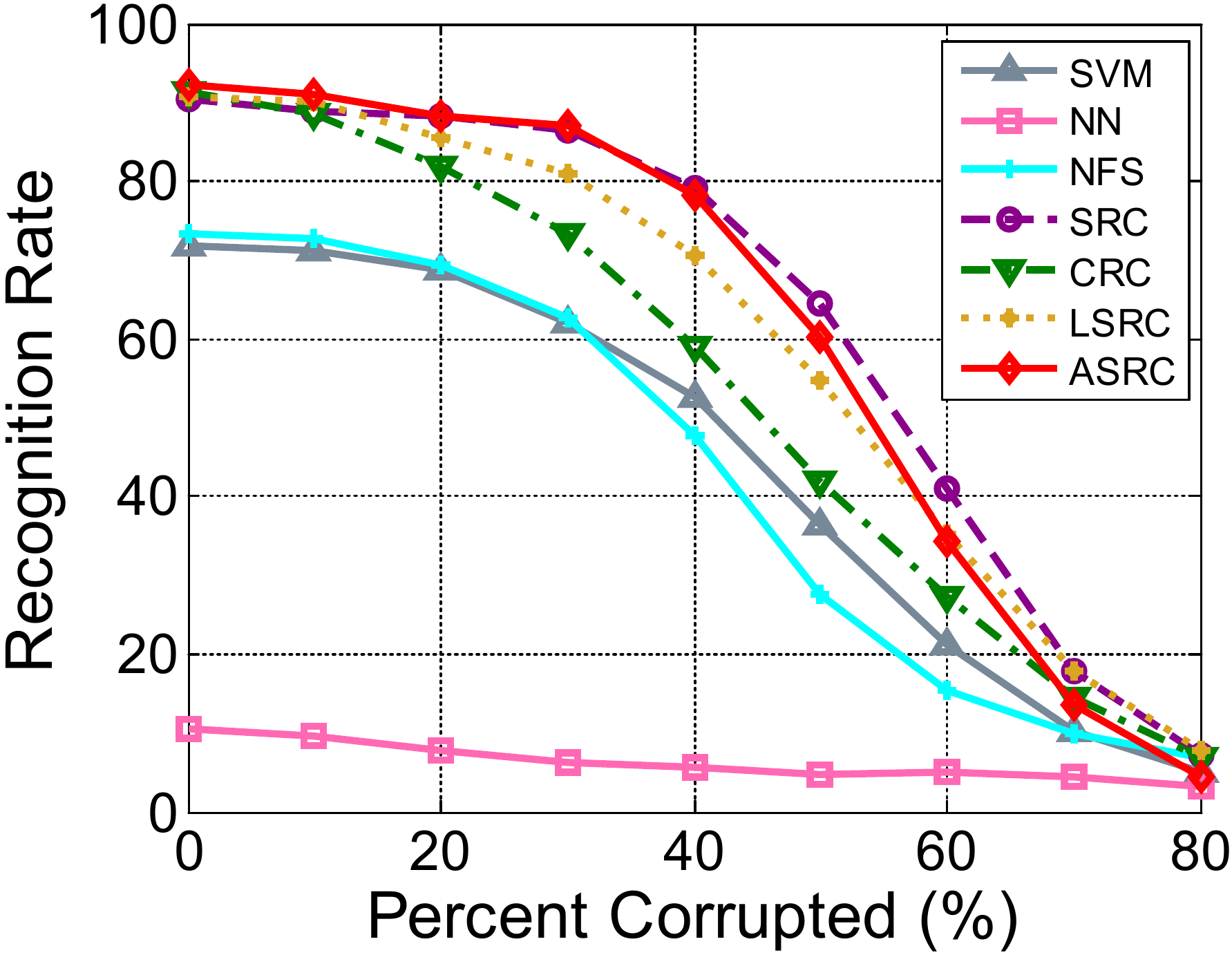}
            \caption{ AR}
        \end{subfigure}
    \caption{The recognition rate across the entire range of corruption for ASRC and its competing algorithms on (a) the Yale database, (b) the ORL database, and (c) the AR database.}
	\vspace{-1em}
\end{figure*}

The results indicate that the correlation information can compensate the corrupted part in the query images. The SRC approach is supposed to be robust to occlusions, as only a fraction of coefficients will be corrupted by the occlusion by using $\ell_1$-minimization. However, the $\ell_1$-minimization requires a large sum of training samples. Thus, SRC obtains a relatively better recognition performance on the AR database with 700 images for training. CRC uses the $\ell_2$-minimization to express occluded images, and most of the coefficients will be corrupted. Thus, CRC is not as robust as SRC and ASRC on two of the three databases (Figure 6b and Figure 6c). We can see that our method outperforms the other algorithms on Yale and ORL databases in most cases. On the AR database, our method performs as well as SRC and better than other methods. This is because the correlation structure suffices the corruption recovery in some ways. Thus,
properly harnessing sparsity and correlation will improve the robustness.

\subsection{Experimental Results on UCI Data Sets}

\begin{table*}[!t]
\centering
\caption{\small{Experimental results on UCI data sets
}}
\label{TabUCI}
\begin{tabular}{ |l |l| l| l| l| l| l| l|}
\hline
 Data Sets &SVM & NN   & NFS  & SRC   & CRC & LSRC   & ASRC
 \\ \hline\hline
 Diabetes  &\textbf{76.96}$\pm$4.97&72.24$\pm$3.90  & 66.05$\pm$3.27   & 65.13$\pm$5.56   & 66.97$\pm$2.74  & 61.08$\pm$3.03 & 68.95$\pm$4.65 \\ 
 Breast  & 70.38$\pm$8.02 & 68.52$\pm$8.24    & 70.37$\pm$0.00    & 69.63$\pm$6.94    & 70.37$\pm$0.00  & 29.31$\pm$5.88   & \textbf{73.70}$\pm$5.64 \\ 
 Breast\_gy & 70.69$\pm$5.88 & 68.89$\pm$9.11    & 72.59$\pm$6.34    & 70.00$\pm$7.50    & \textbf{75.93}$\pm$5.59 & 28.57$\pm$5.78   & 75.56$\pm$5.84 \\ 
 Heart    &85.19$\pm$5.52& 79.26$\pm$6.10    & 81.11$\pm$4.77    & 78.15$\pm$7.70    & 82.59$\pm$5.25 & 77.04$\pm$7.96   & \textbf{85.93}$\pm$6.49 \\ 
 Hearts   &85.19$\pm$5.52 & 74.07$\pm$4.94    & 80.74$\pm$5.18    & 71.11$\pm$6.00    & 84.81$\pm$4.43  & 77.04$\pm$7.96   & \textbf{85.56}$\pm$5.37 \\  
 Cleve   &81.67$\pm$8.58 & 71.72$\pm$7.05    & 77.24$\pm$5.91    & 70.34$\pm$9.91    & 81.03$\pm$10.57 & 80.24$\pm$6.14   & \textbf{82.07}$\pm$9.02 \\ 
 Vote    & \textbf{93.66}$\pm$3.63 & 91.90$\pm$3.01    & 76.67$\pm$4.60    & 75.24$\pm$6.66    & 92.62$\pm$3.63  & 61.41$\pm$6.01  & 93.10$\pm$4.41 \\  
 German   &\textbf{77.00}$\pm$1.94 & 67.90$\pm$3.51    & 70.00$\pm$0.00    & 75.80$\pm$3.61    & 74.60$\pm$2.91  & 28.50$\pm$4.14   & 75.60$\pm$4.95 \\  
 Ionosphere & 64.00$\pm$22.46 & 64.71$\pm$0.00    & 48.53$\pm$7.10    & 90.29$\pm$6.51    & 93.53$\pm$3.62  & 36.00$\pm$22.46  & \textbf{94.41}$\pm$3.24 \\ 
 Spectf   & 80.07$\pm$33.80 & 78.15$\pm$6.8    & 78.52$\pm$7.37    & 68.15$\pm$11.61    & 79.26$\pm$11.61  & 79.77$\pm$34.07   & \textbf{84.07}$\pm$7.21 \\  
 Wdbc    &\textbf{96.66}$\pm$2.85 & 95.36$\pm$2.10    & 88.39$\pm$1.26    & 92.68$\pm$2.14    & 92.32$\pm$2.53 & 62.51$\pm$18.28  & 96.43$\pm$2.06 \\ 
 Air     &95.32$\pm$2.83 & \textbf{96.76}$\pm$3.24    & 91.76$\pm$4.11    & 94.71$\pm$4.56    & 93.24$\pm$4.81  & 97.26$\pm$2.20  & 93.82$\pm$3.52 \\  
 X8D5K   &\textbf{100.00}$\pm$0.00  & \textbf{100.00}$\pm$0.00    & 98.10$\pm$1.20    & 99.90$\pm$0.32    & \textbf{100.00}$\pm$0.00  & \textbf{100.00}$\pm$0.00    & \textbf{100.00}$\pm$0.00 \\ 
 Glass   &37.33$\pm$8.68 & 22.22$\pm$8.28    & 33.89$\pm$8.05    & 30.00$\pm$12.88    & \textbf{38.33}$\pm$5.52 & 33.28$\pm$7.14   & 37.22$\pm$5.89 \\  
 Glass\_gy &38.29$\pm$6.76 & 27.78$\pm$7.86    & 37.78$\pm$9.00    & 32.22$\pm$9.00    & \textbf{40.56}$\pm$9.09 & 26.06$\pm$10.01   & 40.00$\pm$8.20 \\  \hline
\end{tabular}
\end{table*}
In order to test the effectiveness of our algorithm on general pattern recognition problems, we conduct experiments on 15 data sets selected from the UCI repository. The details of the data sets are described in Table 1. On the UCI data sets, we adopt 10-fold cross validation and record the mean and standard deviation of accuracy. The experimental results are shown in Table 2 with the best results highlighted in bold. 

From the results in Table \ref{TabUCI}, we can see that ASRC outperforms NFS on all the data sets. Compared with SRC, our algorithm wins 14 out of 15 times except on the data set German (0.2\%). Compared to NN, ASRC wins 12 times, and ties once on the data set X8D5K. The accuracy rates of CRC are higher than ASRC on three data sets, but the difference is not statistically significant. The performance of LSRC is not stable which limits its applications in general pattern recognition problems. SVM obtains the best results compared with the others for it is a classical classifier, but it is still inferior to our algorithm for 8 times. To sum up, all the experimental results demonstrate that our algorithm is an efficient pattern recognition method.

Experimental results on face image databases validate the effectiveness of our method on face recognition. The correlation information among training images can compensate the pixel occlusion, image misalignment and variations. Meanwhile, the adaptive integration of the dictionary correlation and sparsity also helps handle general pattern recognition problems, which is validated on the UCI data sets.

\section{Conclusion}
\label{sec5}
In this paper, we propose an Adaptive Sparse Representation based Classification (ASRC) method for face recognition. Different from SRC and CRC, ASRC considers both sparsity and correlation in representation. Considering the sparsity, ASRC selects the most discriminative samples for presentation. With the correlation information, ASRC can rectify the occlusion, corruption or variations by training images of other subjects. ASRC can obtain comparable results to SRC when the dictionary is with low correlation, and performs as well as CRC when the data are with high correlation. In other cases, ASRC will obtain an accurate linear representation with the most related and discriminative samples which will guarantee the good recognition performance. Compared with LSRC, our algorithm considers complementary correlation information and is adaptive to the structure of the dictionary. Experimental results on real-world face image data sets show that our method outperforms state-of-the-art face recognition methods, such as SVM, NN, NFS, SRC, CRC and ASRC in terms of recognition precision and robustness. Meanwhile, ASRC can also be treated as a robust classifier and solve other problems such as motion segmentation, activity recognition, subspace learning and so on. The experiments conducted on benchmark data sets from the UCI repository also validate the efficiency of our method in general pattern recognition problems. Thus, our algorithm can be applied not only in face recognition, but also in other important tasks, such feature selection, event detection in multimedia.


%

\section*{Acknowledgment}

 The authors would like to thank the anonymous reviewers for their constructive comments. This work is supported by the National 973 Program of China under grant 2014CB347600, the NSFC under grant nos. 61273292, 61272393, 61322201, the Specialized Research Fund for the Doctoral Program of Higher Education under grant 20130111110011, the
 Program for New Century Excellent Talents in University under grant NCET-12-0836, the Open Project Program of the National
 Laboratory of Pattern Recognition (NLPR), and Singapore Ministry of Education under research Grant MOE2010-T2-1-087. J. Wang would like to thank the China Scholarship Council for their support.

\bibliographystyle{IEEEtran}
\bibliography{ASRCreference}

\end{document}